\documentclass{article}
\usepackage{amssymb}
\usepackage{amsmath}
\usepackage{xcolor}
\usepackage{graphicx}
\usepackage{capt-of}
\usepackage{float}
\usepackage{wrapfig}
\usepackage{lipsum}
\usepackage{subcaption}
\usepackage{multicol}
\usepackage{booktabs}
\usepackage{colortbl}
\usepackage{makecell}
\usepackage{pifont}

\usepackage[preprint]{corl_2025} % Uncomment for pre-prints (e.g., arxiv); This is like ``final'', but will remove the CORL footnote.
\newcommand{\xmark}{\ding{55}}
\newcommand{\todo}[1]{}
\renewcommand{\todo}[1]{{\color{red}{#1}}}

\title{GraspVLA: a Grasping Foundation Model Pre-trained on Billion-scale Synthetic Action Data}

\author{
 \hspace{-1cm}\textbf{Shengliang Deng}$^{\ast, 1, 3}$ \textbf{Mi Yan}$^{\ast, 1, 2}$ \textbf{Songlin Wei}$^{1, 2}$ \textbf{Haixin Ma}$^{1}$ \textbf{Yuxin Yang}$^{1}$
  \textbf{Jiayi Chen}$^{1,2}$ \textbf{Zhiqi Zhang}$^{1,2}$ \\ \hspace{-1.5cm} \textbf{Taoyu Yang}$^{2}$ \quad \textbf{Xuheng Zhang}$^{2}$ \quad \textbf{Wenhao Zhang}$^{2}$ \quad \textbf{Heming Cui}$^{3}$ \quad \textbf{Zhizheng Zhang}$^{\dagger, 1,4}$ \quad \textbf{He Wang}$^{\dagger, 1,2,4}$\\[0.3cm]
  \hspace{-2cm}
  \vspace{-0.5cm}
}

\begin{document}
\maketitle

\begin{center}
    \vspace{-1cm}
   \includegraphics[width=\textwidth]{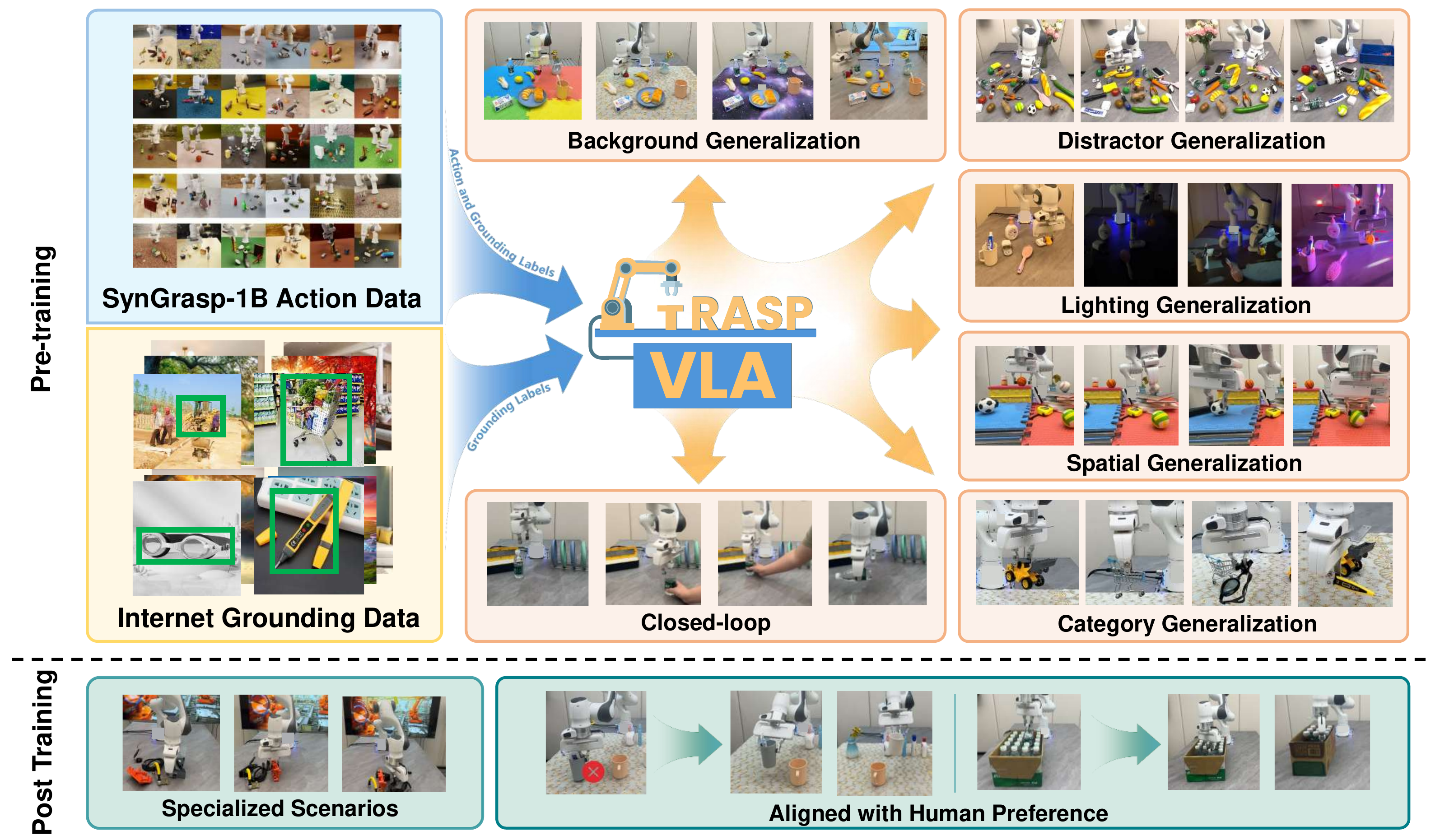}
   \captionof{figure}{GraspVLA is a grasping foundation model pre-trained exclusively on billion-scale synthetic action data and co-trained with Internet semantics data. It exhibits direct sim-to-real transfer and strong zero-shot generalization across diverse aspects, as well as few-shot adaptability to specialized scenarios and human preferences.
   % Pre-trained on billion-scale synthetic action data, GraspVLA demonstrates direct sim-to-real transfer and achieves strong zero-shot generalization in grasping across diverse configurations. Leveraging web-scale grounding data, it can grasp instances from open-vocabulary object categories. After being post-trained with a small number of demonstrations, GraspVLA effectively learns to grasp in specialized scenarios and align with human preference.
   }
   \label{fig:teaser}
   \vspace{-3mm}
\end{center}

\begin{abstract}
\renewcommand*{\thefootnote}{\fnsymbol{footnote}}
\footnotetext{
\hspace*{-1.8em}$^\ast$ Equal contribution with the order determined by rolling dice. $^\dagger$ denotes corresponding authors. \\
Correspondence to \href{zhangzz@galbot.com, hewang@pku.edu.cn}{\texttt{zhangzz@galbot.com, hewang@pku.edu.cn}}.\\
$^1$Galbot, $^2$Peking University, $^3$The University of Hong Kong, $^4$Beijing Academy of Artificial Intelligence
}
Embodied foundation models are gaining increasing attention for their zero-shot generalization, scalability, and adaptability to new tasks through few-shot post-training.
However, existing models rely heavily on real-world data, which is costly and labor-intensive to collect. 
Synthetic data offers a cost-effective alternative, yet its potential remains largely underexplored.
To bridge this gap, we explore the feasibility of training Vision-Language-Action (VLA) models entirely with large-scale synthetic action data.
We curate SynGrasp-1B, a billion-frame robotic grasping dataset generated in simulation with photorealistic rendering and extensive domain randomization.
Building on this, we present GraspVLA, a VLA model pretrained on large-scale synthetic action data as a foundational model for grasping tasks.
GraspVLA integrates autoregressive perception tasks and flow-matching-based action generation into a unified Chain-of-Thought process, enabling joint training on synthetic action data and Internet semantics data. This design helps mitigate sim-to-real gaps and facilitates the transfer of learned actions to a broader range of Internet-covered objects, achieving open-vocabulary generalization in grasping.
Extensive evaluations across real-world and simulation benchmarks demonstrate GraspVLA’s advanced zero-shot generalizability and few-shot adaptability to specific human preferences. 
We will release SynGrasp-1B dataset and pre-trained weights to benefit the community.
Our project page is at \href{https://pku-epic.github.io/GraspVLA-web/}{https://pku-epic.github.io/GraspVLA-web}.
\end{abstract}
\vspace{-4mm}
\vspace{-6mm}
\section{Introduction}
\vspace{-4mm}
The fields of Natural Language Processing (NLP) and Computer Vision (CV) have undergone a paradigm shift with the advent of foundation models. Large-scale models pretrained on vast amounts of Internet data exhibit zero-shot generalization to unseen scenarios \cite{llama,kirillov2023segany, radford2021learningtransferablevisualmodels} and few-shot adaptation for aligning with human preferences \cite{openai2023chatgpt}. Inspired by this success, the foundation model for actions in the physical world has recently been introduced in Vision-Language-Action (VLA) models \cite{rt-2,openvla,pi_0,groot}. These models process robotic visual observations and human instructions to directly generate robot actions. However, unlike vision and language modalities, action data is absent from existing Internet datasets, demanding a new paradigm for data collection.
% The fields of Natural Language Processing (NLP) and Computer Vision (CV) have seen a paradigm shift with the advent of foundation models—large-scale models pretrained on vast amounts of Internet data that exhibit zero-shot generalization to unseen scenarios \cite{llama,kirillov2023segany, radford2021learningtransferablevisualmodels} and few-shot adaptation to align with human preferences \cite{openai2023chatgpt}. Inspried by this, the foundation model for action in the physical world has recently been introduced in RT-2 \cite{rt-2}, giving rise to Vision-Language-Action (VLA) models, which take in robotic visual observations and human instructions to directly output robot actions. However, this new modality, action, requires a new paradigm of data collection since it is absent in existing Internet vision-language data.

Recent research mainly rely on real-world data collection through teleoperation, exemplified by community-driven efforts like Open-X-Embodiment (OXE) \cite{oxe} and DROID \cite{droid} datasets. However, gathering real-world data at a large scale is both labor-intensive and costly, requiring a large number of robots and human operators, as well as diverse physical setups. In contrast, synthetic data offers a more accessible and cost-effective alternative -- yet its potential remains largely underestimated.

To this end, we systematically explore the potential of synthetic data for training VLA models. As a first step in this direction, we focus on grasping, a fundamental robotic manipulation skill. We first curate a billion-frame grasping dataset, SynGrasp-1B, based on advanced ray-tracing rendering \cite{isaac-sim} and physics simulation \cite{mujoco}, marking the first dataset of this scale globally. This dataset incorporates 10,000 unique objects from 240 categories and encompasses extensive domain randomization, ensuring broad coverage of geometric and visual variations.
% To this end, we systematically explore the potential of synthetic data for training VLA models. As a first step in this direction, we focus on grasping, a fundamental robotic manipulation skill. Leveraging advanced ray-tracing rendering \cite{isaac-sim} and physics simulation \cite{mujoco}, we curate a billion-frame grasping dataset, \textbf{SynGrasp-1B}, marking the first dataset of this scale globally. This dataset incorporates 10,000 unique objects from 240 categories and encompasses extensive domain randomization, ensuring broad coverage of geometric and visual variations.

To efficiently learn from this dataset, we propose GraspVLA, an end-to-end network that integrates autoregressive perception tasks and flow-matching-based action generation into a unified Chain-of-Thought (CoT) process, named Progressive Action Generation (PAG). PAG treats perception tasks, i.e., visual grounding and grasping pose prediction, as intermediate steps in action generation, forming a CoT process that causally infers actions. This design enables joint training on synthetic and Internet data in a unified framework, where Internet data is used to train the perception tasks (partial CoT process), and synthetic data is used to train the entire CoT pipeline. Synthetic data provides detailed geometric information about objects for object interactions, while Internet data offers rich semantic knowledge about objects. By leveraging these complementary sources, PAG reduces sim-to-real gaps and facilitates the transfer of learned robotic actions to semantically diverse Internet-covered objects, thereby enabling open-vocabulary grasping.

% To efficiently learn from this dataset, we propose GraspVLA, an end-to-end network that combines an autoregressive vision-language model (VLM) for perception and a flow-matching expert for action generation, inspired by $\pi_0$ and GR00T \cite{pi_0,groot}.

% To scale GraspVLA's grasping capabilities beyond synthetic categories, we introduce a novel Progressive Action Generation (PAG) mechanism. PAG leverages the complementary strengths of synthetic action data and Internet grounding data. Specifically, the VLM is trained on both datasets to generate bounding boxes for diverse objects in a unified prompt format. The VLM is also trained on synthetic data to predict tokenized grasp poses following bounding box generation. Meanwhile, an action expert is trained on synthetic data conditioned on the features derived from the VLM. This design not only extracts rich semantic knowledge from Internet data, but also provides explicit guidance for action generation. PAG also resembles Chain-of-Thought \cite{wei2022chain} that breaks down complex tasks into sub-tasks, thereby greatly enhancing the model's ability of open-vocabulary grasping.

Empowered by our curated billion-scale synthetic grasping dataset and the proposed PAG mechanism, GraspVLA achieves direct sim-to-real generalization and demonstrates impressive zero-shot performance. To the best of our knowledge, this is the first work to reveal the significant potential of synthetic data in training VLA models for manipulation. Extensive experiments conducted in both real-world settings and the LIBERO \cite{liu2024libero} simulation benchmark demonstrate the model's robustness across diverse variations. In addition, GraspVLA shows excellent generalization to long-tail object categories absent from synthetic action data, such as chargers, towels, and swimming goggles. Compared to AnyGrasp \cite{fang2023anygrasp}, the state-of-the-art in traditional grasping detection algorithms, GraspVLA supports natural language instructions and delivers a robust closed-loop grasping policy. It achieves comparable performance on common objects while significantly outperforming AnyGrasp on transparent objects. Moreover, GraspVLA demonstrates strong few-shot adaptability to user preferences in specified application scenarios that extend beyond standard grasping behaviors, such as avoiding contact with the interior of drinking cups to maintain cleanliness and sequentially grasping bottles in densely packed environments.

% Empowered by this large-scale dataset and powerful network, GraspVLA demonstrates direct sim-to-real transfer and achieves superior zero-shot generalizability, revealing the high potential of synthetic data for training VLA models. Extensive experiments in both real world and LIBERO \cite{liu2024libero} simulation benchmark show that our model successfully handles all kinds of variations. Our PAG mechanism further extends GraspVLA’s generalization to long-tail object categories -- such as chargers, towels, and swimming goggles -- which are absent from the synthetic action data. Compared to AnyGrasp \cite{fang2023anygrasp} -- a state-of-the-art method specially designed for grasping, our model achieves comparable performance on common objects and significantly outperforms it on transparent objects. Furthermore, GraspVLA demonstrates strong few-shot adaptability to specific human preferences that go beyond common grasping behavior, such as avoiding touching the inside of a drinking cup to maintain cleanliness, and sequentially grasping bottles in a densely packed environment.

In summary, our contributions are as follows: a) we introduce a novel pretraining paradigm that relies entirely on synthetic action data, significantly reducing the real world action data acquisition burden, b) we curate a billion-frame robotic grasping dataset, SynGrasp-1B, the first dataset of this scale globally, c) we propose Progressive Action Generation to co-train synthetic actions with Internet data, extending GraspVLA’s skills to novel object categories, and d) extensive experiments demonstrate GraspVLA’s foundation capability, including strong zero-shot generalization and efficient few-shot adaptability in real-world.
\vspace{-5mm}
\section{Related Work}
\vspace{-4mm}

\textbf{Vision-Language-Action (VLA) Models.} Recently, a number of works\cite{rt-1,bharadhwaj2023roboagent, hpt, ecot, li2024manipllm, li2024llara, goyal2024rvt, zhen20243d, zhang2024navid} explored training an end-to-end VLA by learning from large-scale demonstration data. RT-2 \cite{rt-2} and OpenVLA \cite{openvla} propose to leverage pre-trained vision-language models (VLMs) \cite{pali-x, palm-e} to exploit the rich knowledge from Internet dataset. Following the success of pre-trained VLMs, several works \cite{octo, pi_0, li2024cogact, groot, wen2024tinyvlafastdataefficientvisionlanguageaction, liu2024rdt} explore leveraging additional action expert to generate multi-modal actions with high fidelity. Others \cite{gr-2, ye2024latent, bharadhwaj2024gen2act, yang2024spatiotemporal, zhao2025cot, tian2024predictiveinversedynamicsmodels} adopt generative pre-training on Internet-scale video data to learn from human videos. However, limited by the scale of real-world robotic data, existing VLA models mainly rely on in-domain post-training for deployment. Concurrent work, $\pi_{0.5}$ \cite{pi0_5}, proposes improving generalization by leveraging multimodal web data and cross-embodiment data, enabling direct out-of-the-box deployment. While our work also targets zero-shot deployment, we take a different approach—exclusively pre-training on large-scale synthetic data—and demonstrate strong zero-shot generalization.

\textbf{Synthetic Data.} With the fast development of GPU-accelerated simulation and photo-realistic rendering, synthetic data generation has become a popular approach to train robotic models. Previous works \cite{bousmalis2017usingsimulationdomainadaptation, acronym, dexnet} pioneered the use of simulated data with domain randomization to train open-loop grasping models. Recently, several works \cite{mimicgen, dexmimicgen, skillmimicgen} explore automatically augmenting human demonstrations in simulation by randomizing object configurations and leveraging motion planning to generate realistic robot trajectories. Another line of work \cite{robosplat, demogen, genaug, roise} synthesizes data from a few human demonstrations utilizing text-to-image generation models and multi-view stereoscopic rendering, without requiring any physical simulation. While these methods \cite{simtoreal} still rely on human demonstrations to generate augmented data, our work explores direct sim-to-real transfer by leveraging large-scale synthetic data together with pre-trained vision and language backbones.

\textbf{Grasping.} Grasping is an essential skill \cite{grasp-survey} for embodied agents and has been actively studied in the past decade. 
Some works tackle this problem through open-loop grasp detection \cite{graspnet, fang2023anygrasp, 6dof-grasp} and then control the end effector using a motion planner. 
Such modular-based systems usually suffer from poor depth perception \cite{wei2024droma} and lack of failure recovery behavior \cite{liu2024efficient, geng2023sage}. 
Another line of research explores vision-based grasping systems in an end-to-end and closed-loop manner, either through reinforcement learning \cite{qt-opt} or imitation learning \cite{song2020grasping}. 
With the advent of vision-language foundation models \cite{llama, alayrac2022flamingo, prismatic-vlm}, several works aim to generalize grasping to open-vocabulary objects \cite{Grasp-Anything, moo, graspgpt, lu2023vl, ding2024open6dor} by building a modular system that combines a grasp detection model with a VLM.
While these methods achieve impressive results in standard grasping, they face challenges in adapting to specialized tasks, such as grasping with specific constraints.
\vspace{-4mm}
\section{\label{datagen}SynGrasp-1B Dataset Generation}
\vspace{-4mm}

\begin{figure}[htbp!]
\centering
\includegraphics[width=\linewidth]{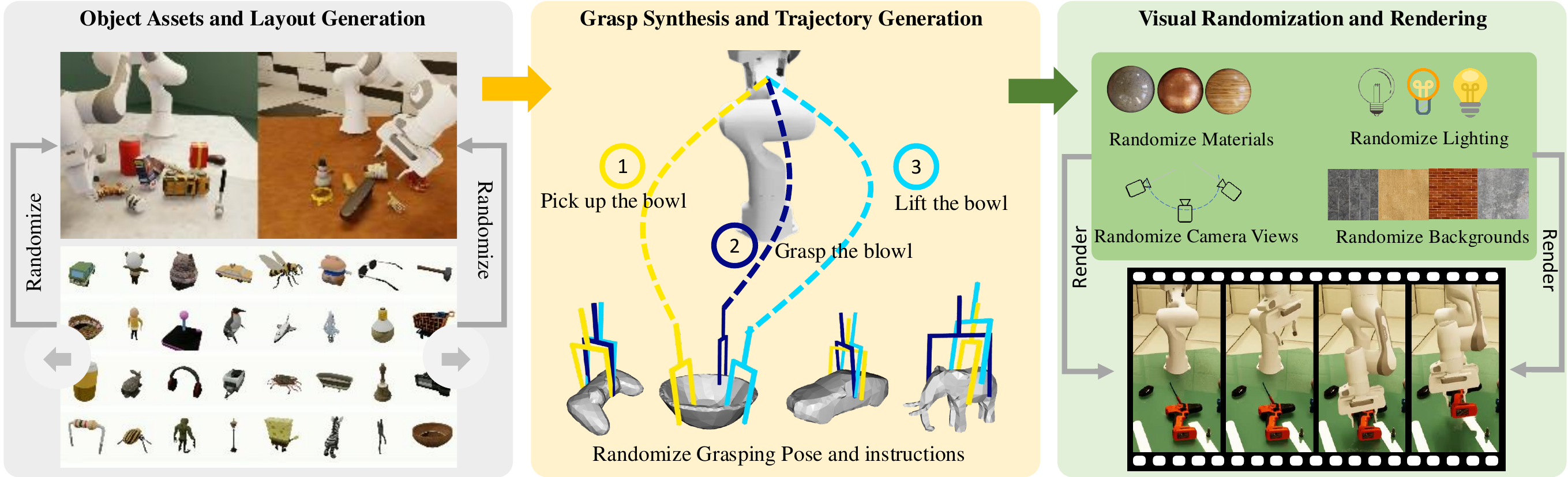}
\caption{\textbf{Data generation pipeline:} We first curated over 10,680 object meshes from Objaverse \cite{objaverse} that are suitable for tabletop grasping and randomly selected and placed these objects on the table (left). Next, we used CuRobo to plan grasping trajectories with randomized grasp poses and instructions (middle). Finally, we applied domain randomization to materials (table and robot), lighting, camera views, and backgrounds to simulate and render the trajectories (right).}
\label{fig:data_pipeline}
\vspace{-5mm}
\end{figure}

Training a generalizable foundation model requires a large-scale dataset encompassing diverse objects and environmental conditions. Instead of relying on costly real-world human data collection, we propose training entirely on synthetic data -- which offers greater diversity at a fraction of the time and expense. We now detail the core components of our synthetic data generation pipeline.

\textbf{Object Assets and Layout Generation.}
We utilize the LVIS subset of the Objaverse dataset \cite{objaverse} and carefully filter out unsuitable categories, such as weapons, resulting in a total of 240 categories and 10,680 instances. We randomly scale these objects and drop them in various poses onto a table, generating diverse and physically plausible scenes. More details can be found in the supplementary.

\textbf{Grasp Synthesis and Trajectory Generation.}
Given initial layouts, we utilize advanced modular system to establish an expert policy for generating high-quality trajectories for grasping and lifting target objects. For each object instance, we leverage grasp synthesis algorithm \cite{bodex} to generate stable antipodal grasps. We then use motion planning algorithms CuRobo \cite{curobo} to plan collision-free trajectories to reach the open-loop grasp pose and lift the object. We validate all candidate trajectories in the MuJoCo physics simulator \cite{mujoco} to ensure successful lifting of the object.

\textbf{Visual Randomization and Rendering.}
Given diverse layouts and corresponding trajectories, we render high-quality RGB images with randomized lighting, backgrounds, and camera settings using Isaac Sim \cite{isaac}, which offers efficient photo-realistic ray-traced rendering. We employ various light sources with extensive randomization, including point, directional, and dome lights. Images are rendered from two different viewpoints to provide a comprehensive view of the scene, with randomized extrinsics around predefined centers. More details are provided in the supplementary material.

We further highlight two major considerations in the design of our data generation pipeline:

\textbf{Efficient Data Generation.}
We develop three key strategies to improve the efficiency. 
High-quality meshes are often large, leading to lengthy loading times and significant memory usage. We implement a caching mechanism to avoid redundant loading while ensuring data diversity.
Second, we implement asynchronous data writing, allowing images and labels to be saved in parallel, thereby improving overall efficiency in data generation.
Finally, we employ parallel physics simulation and rendering to further improve efficiency. 
Please refer to the supplementary for more details.

\textbf{Tailoring Data for Imitation Learning.}
To ease the difficulty of imitation learning, we introduce two improvements. First, while open-loop grasping \cite{fang2023anygrasp} employs a two-step process (pregrasp positioning followed by grasp execution) to avoid collision, this segmented approach creates pauses. Imitation policies trained on such data often exhibit hesitation \cite{openvla, dalal2023imitating}. Instead, we implement single-step motion planning, prioritizing trajectory smoothness over planning success rates. Second, we introduce randomized initial robot poses to improve workspace exploration and observation diversity in expert demonstrations, enhancing model robustness \cite{datascalinglaw}.

With this pipeline, we generate our billion-frame dataset, SynGrasp-1B, using 160 NVIDIA 4090 GPUs for 10 days. We provide data diversity analysis in the supplementary.

\vspace{-4mm}
\section{\label{model}Model}
\vspace{-4mm}

% \textbf{Problem Formulation.}
% The formulation of the language-conditioned close-loop grasping in this work is as follows: in each episode, the model is given a language instruction $l$ in the form of ``pick up \{object\_name\}'', and the instruction remains constant at all the time steps.
% At each time step $t$, the model could access observations $o_i (i\le t)$ at current time step and all previous time steps, and generates action $a_t$ to control the end effector of the robot arm.
% Observations include two RGB images from front and side views, along with 7-DoF pose $p$ of the end effector.
% The proprioception consists of the 3D position and Euler angles of the end effector in the robot base coordinate, and the gripper's open/close state, i.e., $p = (x, y, z, r_x, r_y, r_y, openness)$.
% The action is in the form of $(\Delta x, \Delta y, \Delta z, \Delta r_x, \Delta r_y, \Delta r_z, openness)$, similar to previous work \cite{rt-2,octo,openvla}.
% To reduce computational costs, we utilize RGB images only from the current time step.
% We incorporate proprioception from both the current and previous time steps such that the model could leverage previous motion for generating smooth actions.

\begin{figure*}[h]
\centering
\includegraphics[width=0.9\textwidth]{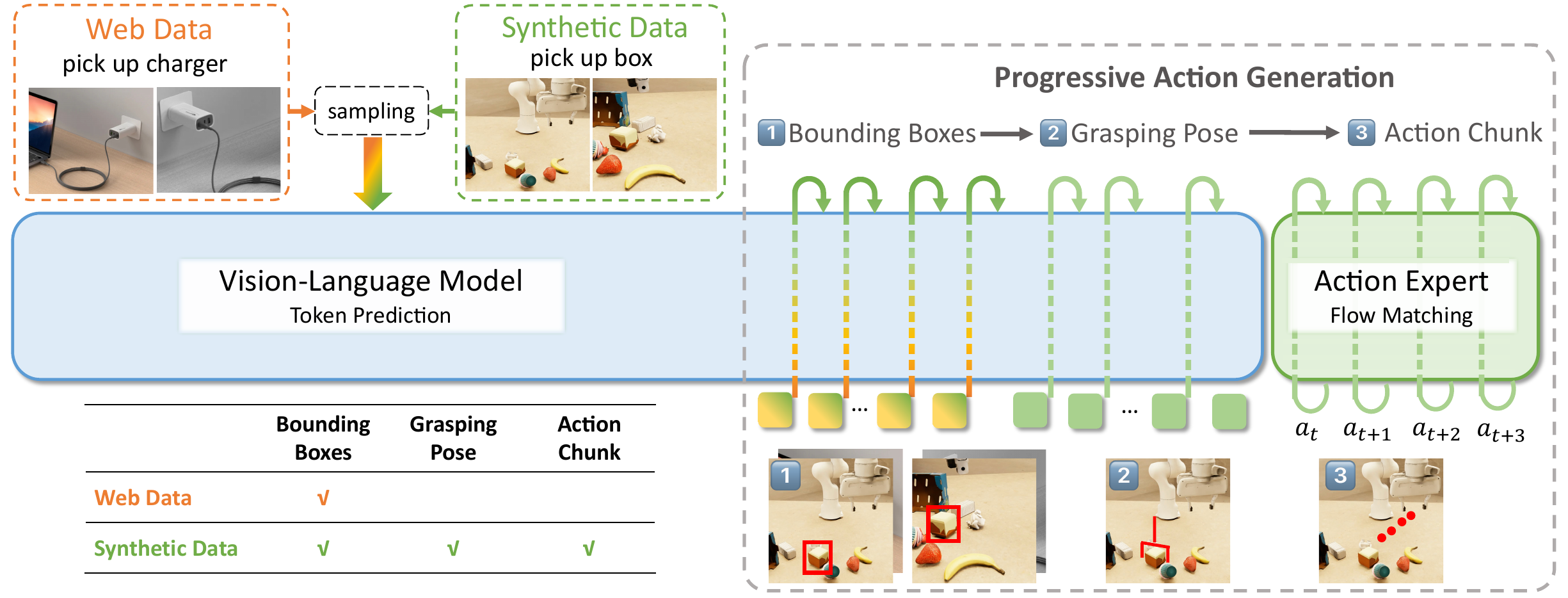}
\caption{\textbf{GraspVLA} consists of an autoregressive vision-language backbone and a flow-matching based action expert. It exploits the synergy between Internet grounding data and synthetic action data with a Progressive Action Generation mechanism: the model first predicts 2D bounding boxes of the target object for both synthetic data and web data, and additionally generates grasp pose and chunked actions for synthetic data.}
\label{fig:model_pipeline}
\vspace{-4mm}
\end{figure*}

\textbf{Overall Architecture.}
GraspVLA integrates a Vision-Language Model (VLM) with an action expert \cite{pi_0}, connected through a Progressive Action Generation (PAG) mechanism, as illustrated in Figure \ref{fig:model_pipeline}.
The VLM takes observation images and a text instruction for vision-language joint perception.
It comprises a trainable large language model (InternLM2 1.8B \cite{internlm2}), a vision encoder that fuses features from frozen DINO-v2 \cite{dinov2} and SigLIP \cite{siglip} inspired by OpenVLA \cite{openvla}, and a trainable projector from the vision space to the language space.
We use a conditional flow matching action expert \cite{lipman2022flowmatching} for fine-grained end-effector action generation.
We further introduce PAG to efficiently transfer knowledge learned from Internet grounding dataset to grasping skills.

% \todo{Following OpenVLA\cite{openvla}, we adopt a transformer-based architecture for our model and generate outputs in an autoregressive manner.
% Illustrated in Figure \ref{fig:model_pipeline}, the model comprises a vision encoder that fuses features from DINO-v2 and SigLIP, a trainable projector, and a trainable Large Language Model (LLM) based decoder. At each time step, the vision encoder encodes the current RGB images of front and side view, and the results of two views are concatenated along the sequence length dimension as the model inputs.
% We adopt a learnable projector to convert vision tokens into input tokens of LLM for a better alignment with the language modality.
% Thereafter, the language instruction is tokenized and appended to the image tokens. As for the decoding part, we employ an open-source LLM, and train it to predict robot actions at each time step in its output space, by discretizing robotic actions into a set of action tokens. The discretization follows the common practices in prior VLA works \cite{rt-2,octo,openvla}.}

\textbf{Progressive Action Generation.}
While GraspVLA learns generalizable grasping skills from our SynGrasp-1B dataset, it is constrained by the set of categories present in the synthetic dataset.
To scale the grasping policy to novel categories, a straight-forward approach is to co-train with Internet grounding dataset as separate tasks, and rely on the model to implicitly generalize to object categories learned from the grounding dataset.

Alternatively, we formulate image grounding and grasp pose prediction as intermediate steps to generate action. Specifically, the VLM is trained to generate 2D bounding boxes for both Internet grounding dataset and synthetic action dataset in a unified format. Then, for the synthetic dataset, the VLM further predicts the target grasp pose in the robot's base frame.
Finally, the action expert generates action chunk conditioned on the VLM's key-value cache of both input and intermediate reasoning tokens.
To facilitate accurate 3D sensing, the proprioceptions from the latest two timesteps are tokenized and inserted before generating grasp pose.
To align the Internet dataset with the dual-camera setup of SynGrasp-1B, input images are duplicated to match the number of views and independently augmented with random resizing, cropping, horizontal flipping, and color jittering.
Both datasets share the same text prompt template, generating bounding box tokens first. This unified training strategy exploits the synergy between the Internet grounding and synthetic datasets, and resembles the Chain-of-Thought reasoning mechanism widely studied and proven as an effective measure to handle highly complex tasks in large language models \cite{wei2022chainofthought}.

% The resulting training sample for grasping actions is as follows:
% \begin{center}
% \fcolorbox{black}{gray!10}{\parbox{.9\linewidth}{\{visual tokens\} Q: What should the robot do to pick up \{object description\}? A: \{bounding box\}\{proprioception\}\{grasp pose\}}}
% \end{center}

\textbf{Joint Training of VLM and action expert.}
In each batch, we randomly sample from the Internet dataset (GRIT \cite{Kosmos2}) and the synthetic action dataset.
The former is used solely to supervise the VLM's bounding box prediction in an auto-regressive manner.
The latter supervises bounding box, grasp pose, and flow-matching-based action prediction.
The loss of VLM is formally defined as:
\[
\mathcal{L}_{\text{S2}} = - \sum_{n=1}^{N_{\text{bbox}}} \log P_\theta (\mathbf{y}_{\text{bbox}, n} \mid \mathbf{x}, \mathbf{y}_{\text{bbox}, <n}) \ - \mathbf{1}_{\text{synthetic}} \cdot \sum_{n=1}^{N_{\text{grasp}}} \log P_\theta (\mathbf{y}_{\text{grasp}, n} \mid \mathbf{x}, \mathbf{y}_{\text{bbox}}, \mathbf{y}_{\text{grasp}, <n}),
\]

% \[
% \begin{aligned}
% \mathcal{L}_{\text{S2}} = & \ \mathbb{E}_{(\mathbf{x}, \mathbf{y}) \sim \mathcal{D}} \left[ - \sum_{t=1}^{T_{\text{bbox}}} \log P_\theta (\mathbf{y}_{\text{bbox}, t} \mid \mathbf{x}, \mathbf{y}_{\text{bbox}, <t}) \right] \\
% & + \mathbb{E}_{(\mathbf{x}, \mathbf{y}) \sim \mathcal{D}_{\text{robotic}}} \left[ - \sum_{t=1}^{T_{\text{grasp}}} \log P_\theta (\mathbf{y}_{\text{grasp}, t} \mid \mathbf{x}, \mathbf{y}_{\text{bbox}}, \mathbf{y}_{\text{grasp}, <t}) \right],
% \end{aligned}
% \]
where \(N_{\text{bbox}}\) and \(N_{\text{grasp}}\) are the lengths of the bounding box and grasp pose token sequences respectively, \(\mathbf{y}_{\text{bbox}, n}\) and \(\mathbf{y}_{\text{grasp}, n}\) are tokens at position \(n\) in their respective sequences,
and \(\mathbf{x}\) is the input images and text.
The action expert is supervised with flow matching loss on chunked end-effector delta actions:
\[
\mathcal{L}_{\text{S1}} = \| v_t(\mathbf{A}_{t}, \mathbf{x}, \mathbf{y}_{\text{bbox}}, \mathbf{y}_{\text{grasp}}) - u_t(\mathbf{A}_{t} \mid \mathbf{A}_{0}) \|^2,
\]
% \[
% \mathcal{L}_{\text{S1}} = \mathbb{E}_{(\mathbf{x}, \mathbf{y}) \sim \mathcal{D}_{\text{robotic}}} \left[ \sum_{h=1}^H \| v_t(\mathbf{a}_{h, t}; \theta) - u_t(\mathbf{a}_{h, t} \mid \mathbf{a}_{h, 0}) \|^2 \right],
% \]
where $t\in[0,1]$ is the flow matching timestep, \( \mathbf{A}_{t} \) is the noised action trunk at \( t \),
\( v_t(\cdot) \) is the model predicted flow matching vector field,
\( u_t(\mathbf{A}_{t} \mid \mathbf{A}_{0}) \) is the ground-truth vector field.
We empirically found a simple sum of $\mathcal{L}_{\text{S2}}$ and $\mathcal{L}_{\text{S1}}$ for the overall loss yields good performance.
\vspace{-4mm}
\section{Experiments}
\vspace{-4mm}

We evaluate GraspVLA to answer the following questions: (1) How does GraspVLA compare with existing work under various generalization factors? (2) How does GraspVLA scale with the amount of data? (3) How much do our design choices contribute to GraspVLA's performance? (4) How well does GraspVLA support few-shot post-training for specialized preferences?

% \vspace{-3mm}
\subsection{Zero-Shot Comparison with VLAs in Real World}
\label{sec:exp_setup}

\vspace{-\baselineskip}
\begin{table}[htbp]
% \vspace{-2mm}
\centering
\resizebox{0.98\columnwidth}{!}{
    \begin{tabular}{lcccccccccccc}
    \toprule
     & \multicolumn{6}{c}{\textbf{Synthetic Categories}} & \multicolumn{6}{c}{\textbf{Web Categories}}\\
     \cmidrule(lr){2-7} \cmidrule(lr){8-13}
    & basic$\uparrow$ & light$\uparrow$ & b.g.$\uparrow$ & dis.$\uparrow$ & height$\uparrow$ & SPL$\uparrow$ & basic$\uparrow$ & light$\uparrow$ & b.g.$\uparrow$ & dis.$\uparrow$ & height$\uparrow$ & SPL$\uparrow$\\
    \cmidrule(lr){2-6} \cmidrule(lr){7-7} \cmidrule(lr){8-12}  \cmidrule(lr){13-13}
    Diffusion Policy \cite{dp} & 30.0 & 16.6 & 16.6 & 13.3 & 13.3 & 12.3 & - & - & - & - & - & - \\
    Octo \cite{octo} & 16.6 & 3.3 & 0.0 & 0.0 & 3.3 & 3.2 & 0.0 & 3.3 & 0.0 & 0.0 & 0.0 & 0.4 \\
    OpenVLA \cite{openvla} & 20.0 & 13.3 & 16.6 & 0.0 & 13.3 & 8.8 & 3.3 & 6.6 & 13.3 & 0.0 & 6.6 & 4.1 \\
    $\pi_0$(w/ $\pi_0$ pre-train)\cite{pi_0} & 66.6 & 63.3 & 60.0 & 60.0 & 56.6 & 42.3 & 33.3 & 36.6 & 30.0 & 26.6 & 26.6 & 17.8 \\
    $\pi_0$(w/o $\pi_0$ pre-train)\cite{pi_0}& 80.0 & 76.6 & 80.0 & 86.6 & 76.6 & 51.8 & 40.0 & 40.0 & 36.6 & 36.6 & 33.3 & 36.9 \\
    Ours & \textbf{93.3} & \textbf{96.6} & \textbf{93.3} & \textbf{93.3} & \textbf{90.0} & \textbf{87.2} & \textbf{93.3} & \textbf{90.0} & \textbf{93.3} & \textbf{86.6} & \textbf{86.6} & \textbf{84.7} \\
    \bottomrule
    \end{tabular}
}
\caption{\textbf{Zero-shot comparisons in real-world.}  We compare our method against state-of-the-art imitation learning specialists and large VLA models. All models are fine-tuned on SynGrasp-1B dataset. Our approach achieves the highest grasping success rate on items from both synthetic and web categories using short trajectories. Detailed description of setups is provided in Section \ref{sec:exp_setup}.}
\label{table:main_values}
\vspace{-8mm}
\end{table}

\textbf{Task Definition.}
To evaluate the effectiveness of PAG, we use two groups of objects: \textbf{synthetic categories} and \textbf{web categories}. We define synthetic categories as those present in our SynGrasp-1B dataset, while web categories refer to those exclusively present in Internet grounding dataset.

\begin{wrapfigure}{r}{0.6\textwidth}
    \centering
    \vspace{-\baselineskip}
    \includegraphics[width=\linewidth]{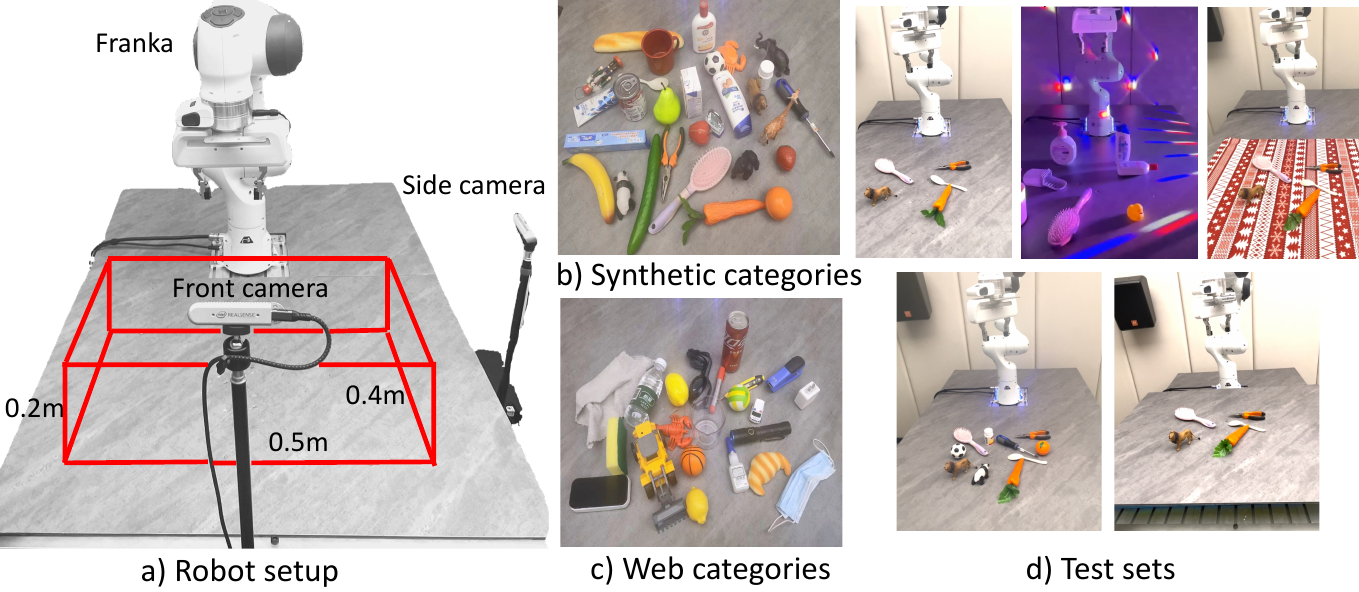}
    \caption{We show our real-world setup in (a), objects used in experiments in (b,c), and 5 test sets corresponding to basic, light, background, distractor, and height settings in (d).}
    \label{fig:setup}
    \vspace{-\baselineskip}
\end{wrapfigure}

For each group of objects, we design 5 test sets: \textbf{basic}, \textbf{lighting}, \textbf{background}, \textbf{distractors}, and \textbf{height}. Each test set contains 15 objects from distinct categories randomly sampled from each group, with 2 trials per object. In other words, we test each method for $15 \times 2 \times 5 \times 2 = 300$ trials in total. We use a disco light to generate different lighting conditions. For background generalization, three distinct tablecloths are selected and interchanged. For distractor generalization, we randomly place 5 additional objects on the table as distractors. For height generalization, we increase the height of the workspace surface by 10 cm. We utilize a Franka Panda arm and employ two Intel RealSense cameras as front and side cameras. The workspace is confined to a 40 cm × 50 cm x 20 cm area in front of the robot. The initial robot and object states are fixed within each trial to ensure fair comparison.

\textbf{Metrics.}
The success rate is defined as the percentage of trials in which the model successfully grasps the target object within 3 attempts. For each object group, we also report the average Success weighted by Path Length (SPL) \cite{anderson2018evaluationembodiednavigationagents}, a widely used metric that weights success rate with motion efficiency by penalizing unnecessarily long paths. It is computed as: $\frac{1}{N} \sum_{i=1}^{N} S_i \frac{l_i}{max(p_i, l_i)},$
where $S_i$ is a binary indicator of success (1 if successful), $l_i$ is the shortest path length achieved by any method in the trial, $p_i$ is the path length taken by the model, and $N$ is the total number of trials.

\textbf{Baselines.}
We compare with multiple baselines including both VLA generalists and imitation learning specialists. For generalists, we use $\pi_0$ \cite{pi_0}, OpenVLA \cite{openvla}, and Octo \cite{octo}, three transformer-based policies pre-trained on large-scale real-world datasets. To ensure fair comparison, we fine-tune all three models on our SynGrasp-1B dataset. Additionally, to assess the effectiveness of pre-training on \textbf{SynGrasp-1B}, we report results of direct fine-tuning $\pi_0$ from its VLM weights \cite{beyer2024paligemma}, without its cross-embodiment robotic pre-training. For specialists, we use Diffusion Policy \cite{dp}, a strong diffusion baseline for visual-conditioned imitation learning. As it lacks language conditioning, we train and test it using only the elephant category. Additional details are provided in the supplementary.

\textbf{Comparisons.} As illustrated in Table \ref{table:main_values}, GraspVLA achieves around 90\% on all test sets and significantly outperforms all baselines, demonstrating strong zero-shot generalizability. Notably, GraspVLA achieves comparable results in both synthetic and web categories, underscoring the effectiveness of PAG. Additionally, the SPL metric reveals that GraspVLA grasps objects with shorter path lengths compared to $\pi_0$ baselines which often exhibit hesitation. Interestingly, the $\pi_0$ baseline without cross-embodiment pre-training performs better than its pre-trained counterpart, suggesting that cross-embodiment pre-training may not be optimal for this specific grasping task on the given robotic arm. We provide failure analysis in the supplementary.

\vspace{-3mm}
\subsection{Zero-Shot Comparison with VLAs in LIBERO Benchmark}
\label{sec:LIBERO}
\vspace{-3mm}
% \begin{wraptable}{r}{0.45\textwidth}
% \vspace{-4mm}
% \resizebox{0.45\columnwidth}{!}{
% \begin{tabular}{lccc}
% \toprule
%  & \textbf{Long} & \textbf{Goal} & \textbf{Object} \\
% \cmidrule(lr){2-4}
% \rowcolor[gray]{0.85}
% \textit{OpenVLA (fine-tuned)} & \textit{70.9} & \textit{78.6} & \textit{91.2} \\
% \rowcolor[gray]{0.85}
% \textit{$\pi_0$ (fine-tuned)} & \textit{\textbf{88.7}} & \textit{\textbf{95.4}} & \textit{\textbf{98.4}}\\ 
% \midrule
% $\pi_0$ (zero-shot) & 0.0 & 0.0 & 0.0\\
% Ours (zero-shot) & \textbf{83.7} & \textbf{93.1} & \textbf{93.9}\\
% \bottomrule
% \end{tabular}
% }
% \caption{\textbf{Comparisons with baselines in LIBERO.}
% }
% \label{table:main_sim}
% \vspace{-3mm}
% \end{wraptable} 

\begin{wraptable}{r}{0.45\textwidth}
\vspace{-4mm}
\resizebox{0.45\columnwidth}{!}{
\begin{tabular}{lccc}
\toprule
 & \textbf{Long} & \textbf{Goal} & \textbf{Object} \\
\cmidrule(lr){2-4}
\rowcolor[gray]{0.85}
\textit{OpenVLA (fine-tuned)} & \textit{33.7} & \textit{56.6} & \textit{65.4} \\
\rowcolor[gray]{0.85}
\textit{$\pi_0$ (fine-tuned)} & \textit{62.7} & \textit{79.4} & \textit{93.8}\\ 
Ours (zero-shot) & \textbf{82.0} & \textbf{91.2} & \textbf{94.1}\\
\bottomrule
\end{tabular}
}
\caption{\textbf{Comparisons with baselines in LIBERO.} The zero-shot performance of GraspVLA surpasses the fine-tuned performance of strong baselines $\pi_0$ and OpenVLA.} 
\label{table:libero}
\vspace{-4mm}
\end{wraptable} 

\textbf{Setup.}
LIBERO \cite{liu2024libero} is a widely used simulation benchmark for robotic manipulation, encompassing diverse tasks and object categories. We evaluate on three LIBERO suites (Long, Goal, Object), excluding Spatial, as its focus on spatial reasoning falls outside our scope. To concentrate on grasping capabilities, we omit non-prehensile tasks (e.g., `turn on the stove') and reformulate task captions as `pick up \{object\}', selecting 7-10 tasks per suite. In line with standard evaluation protocols, each task is rigorously tested with 50 randomized initial configurations, resulting in 350-500 trials per suite. More details are provided in the supplementary.

\textbf{Comparisons.}
As shown in Table \ref{table:libero}, GraspVLA demonstrates satisfactory performance when zero-shot evaluated on LIBERO. It surpasses $\pi_0$ and OpenVLA fine-tuned on the LIBERO dataset, demonstrating strong generalizability. We also observe that the format of task captions significantly affects the performance of fine-tuned models and provide detailed results in the supplementary.

\vspace{-3mm}
\subsection{Zero-Shot Comparison with AnyGrasp in Real World}
\vspace{-3mm}
\begin{wraptable}{r}{0.55\textwidth}
\vspace{-4mm}
\resizebox{0.55\columnwidth}{!}{
\begin{tabular}{lccccc}
    \toprule
     & \multicolumn{2}{c}{\textbf{Language-Conditioned}} & \multicolumn{2}{c}{\textbf{Arbitary Grasping}} & \textbf{Speed} \\
    \cmidrule(lr){2-3} \cmidrule(lr){4-5} 
    & overall & grasp & common & transparent &  \\
    \cmidrule(lr){2-3} \cmidrule(lr){4-5} \cmidrule(lr){6-6}
    AnyGrasp & 91.6 & \textbf{96.6} & \textbf{100.0} & 10.0 & 37 Hz \\
    Ours & \textbf{93.3} & 93.3 & 93.3 & \textbf{86.6} & 5 Hz \\
    \bottomrule
    \end{tabular}
}
\caption{\textbf{Comparison with AnyGrasp.} GraspVLA performs consistently well in both language-guided and arbitrary grasping tasks. In contrast, AnyGrasp is faster and excels at grasping common objects but struggles with transparent objects.
}
\label{table:any_grasp}
\vspace{-5mm}
\end{wraptable} 
\textbf{Setup.} We benchmark GraspVLA against AnyGrasp \cite{fang2023anygrasp}, a state-of-the-art grasp detection model specialized in grasping. For language-conditioned grasping, we integrate AnyGrasp with Grounding DINO \cite{liu2024groundingdinomarryingdino}, a popular open-vocabulary object detector, to filter grasp candidates. We use the same two basic test sets (Section \ref{sec:exp_setup}), with metrics including overall success rate (task completion) and grasping success rate (grasping any object). To isolate grasping performance, we design two additional test sets (30 trials each): one with common household objects and another with transparent objects, where the robot can grasp any object in the scene.

\textbf{Comparisons.} In the language-conditioned test set, both model achieve similar performance, with GraspVLA slightly outperforming AnyGrasp in grounding ability, due to its comprehensive multi-view observation. In arbitrary object grasping, while AnyGrasp achieves a 100\% success rate in grasping common objects, it struggles with transparent objects due to inaccurate depth sensing and incomplete point cloud data. In contrast, GraspVLA maintains consistent performance across both test sets, highlighting its robustness to material variations. However, GraspVLA’s inference speed is significantly slower than AnyGrasp’s, a limitation tied to its large vision-language backbone.

\vspace{-3mm}
\subsection{Scaling Law}
\vspace{-3mm}

\begin{wrapfigure}{r}{0.4\textwidth}
\centering
\vspace{-8mm}
\includegraphics[width=\linewidth]{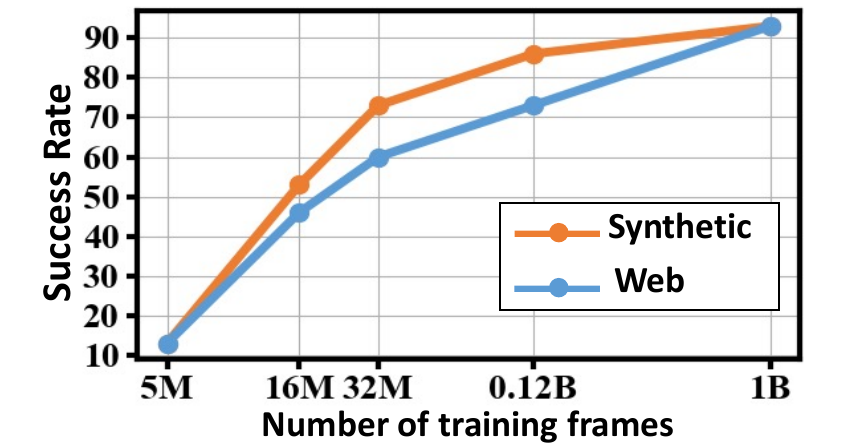}
\caption{The performance scales with the number of training frames, especially for web categories.}
\label{fig:scaling_frames}
\vspace{-0.3cm}
\end{wrapfigure}

Figure \ref{fig:scaling_frames} shows the scaling curve regarding the number of training frames in real world. We observe that the performance improves steadily with the number of training frames and the performance on web categories scales slower than that of synthetic categories, indicating that more training frames are needed for good generalization on web categories. For scaling law regarding the number of training categories and the number of instances per category, please refer to the supplementary.

\vspace{-3mm}
\subsection{Efficient Post-Training}
\vspace{-3mm}
A defining characteristic of foundation models is their ability to adapt to new tasks. To this end, we design three downstream tasks: i) Task 1 -- grasping rare industrial components, ii) Task 2 -- grasping a mug without touching its interior to maintain cleanliness, and iii) Task 3 -- sequential grasping in a densely packed environment. These tasks rigorously benchmark the model’s adaptability capability to three critical challenges: (i) generalizing to new vocabularies, (ii) executing task-specific grasp specifications, and (iii) grasping in order. We collect 100 demos for Tasks 1--2 and 10 per bottle for Task 3. We conduct 10 trials per task and report the overall success rate (task completion) and the grasping success rate (grasping any object).

\begin{figure}[htbp]
\centering
\includegraphics[width=0.9\linewidth]{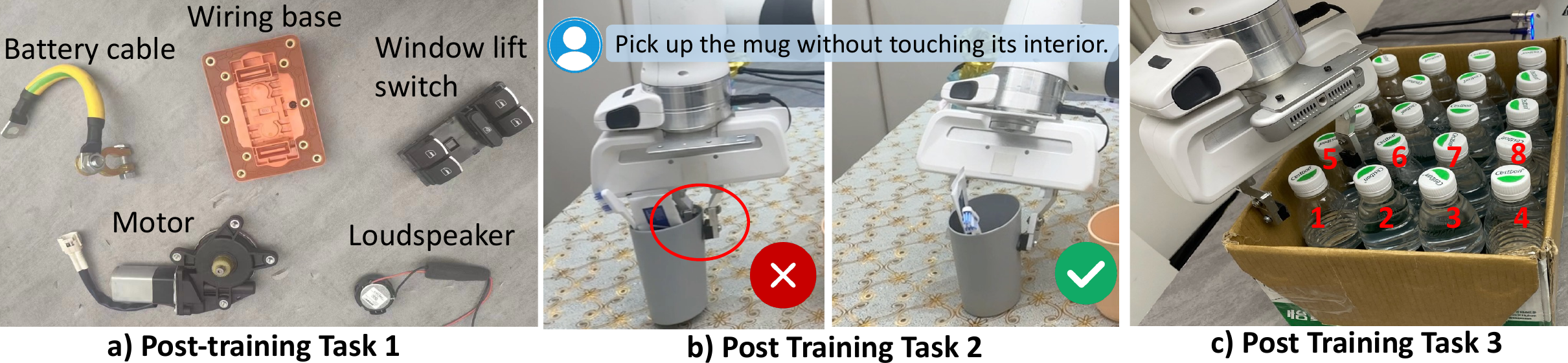}
\caption{\textbf{Real-world post-training.} We experimented with three different post-training tasks to showcase that our model can quickly learn to grasp new items in (a), new grasping patterns in (b), and new grasping behavior in (c).}
\label{fig:post-train}
\vspace{-3mm}
\end{figure}

\begin{wraptable}{r}{0.63\textwidth}
\centering
\vspace{-4mm}
\resizebox{0.63\columnwidth}{!}{
\begin{tabular}{lcccccccc}
\toprule
 & \multicolumn{2}{c}{\textbf{Training Data}} & \multicolumn{2}{c}{\textbf{Task 1}} & \multicolumn{2}{c}{\textbf{Task 2}} & \multicolumn{2}{c}{\textbf{Task 3}} \\ 
\cmidrule(lr){2-3} \cmidrule(lr){4-5} \cmidrule(lr){6-7} \cmidrule(lr){8-9}
& BBox & traj. & overall & grasp & overall & grasp & overall & grasp \\
\cmidrule(lr){2-3} \cmidrule(lr){4-5} \cmidrule(lr){6-7} \cmidrule(lr){8-9}
OpenVLA & - & - & 0 & 0 & 0 & 20 & 0 & 0 \\
$\pi_0$ & - & - & 10 & 20 & 0 & 30 & 0 & 0 \\
Ours & - & - & 40 & 90 & 0 & 80 & 0 & 20 \\ 
\hline
DP & - & \checkmark & - & - & 20 & 60 & 10 & 30 \\
OpenVLA & - & \checkmark & 0 & 0 & 20 & 30 & 0 & 20 \\
$\pi_0$ & - & \checkmark & 60 & 80 & 60 & 70 & 50 & 60 \\
Ours & \checkmark & - & \textbf{90} & \textbf{100} & - & - & - & - \\
Ours(scratch) & \checkmark & \checkmark & 10 & 30 & 10 & 30 & 0 & 20 \\
Ours & \checkmark & \checkmark & \textbf{90} & \textbf{100} & \textbf{80} & \textbf{90} & \textbf{90} & \textbf{90} \\
\bottomrule
\end{tabular}
}
\caption{\textbf{Efficient post-training.} GraspVLA shows superior adaptability to novel tasks, surpassing the model without pre-training and all baselines.}
\label{table:post_train}
\vspace{-3mm}
\end{wraptable} 
As shown in Table \ref{table:post_train}, GraspVLA achieves a 90\% success rate with only bounding box annotations in Task 1, surpassing baselines trained on full action data. This suggests that extending GraspVLA to new objects does not necessitate action annotations, thereby greatly reducing data collection effort. As shown by the last two rows, training from scratch yields lower performance, underscoring the value of our synthetic pre-training. Notably, in Task 3’s dense sequential grasping, GraspVLA learns to avoid collisions with surrounding objects effectively.

\vspace{-3mm}
\subsection{\label{sec:ablation}Effectiveness of Design Choices}
\vspace{-3mm}

\begin{wraptable}{r}{0.35\textwidth}
\centering
\vspace{-4mm}
\resizebox{0.35\columnwidth}{!}{
\begin{tabular}{lcccc}
\hline
 & \multicolumn{2}{c}{\textbf{Synthetic}} & \multicolumn{2}{c}{\textbf{Web}} \\ 
\cmidrule(lr){2-3} \cmidrule(lr){4-5}
 & SR  & SPL  & SR & SPL  \\
\cmidrule(lr){2-3} \cmidrule(lr){4-5}
vanilla & 66.6 & 39.3 & 53.3  & 27.7  \\
+ PAG-2D & 80.0 & 59.2 & 76.7 & 48.9 \\
+ PAG-3D & \textbf{93.3} & \textbf{90.2} & \textbf{93.3}   & \textbf{91.7}  \\ \hline
\end{tabular}
}
\caption{We give a detailed ablation study of our models. With all the design choices enabled the performance boosts significantly.}
\label{table:ablation-model}
\vspace{-3mm}
\end{wraptable} 

As shown in Table \ref{table:ablation-model}, we evaluate the effectiveness of our key design choices using both success rate and SPL metrics on the basic test set described in Sec. \ref{sec:exp_setup}. The vanilla baseline, which employs co-training with Internet grounding data but excludes PAG, serves as our starting point. Introducing 2D bounding boxes as intermediate action steps (PAG-2D) yields significant improvements for web categories. Further enhancement comes with grasp pose prediction (PAG-3D), which substantially reduces hesitation behavior and improves grasping accuracy. This leads to fewer attempts and shorter trajectories, as reflected in the higher SPL scores. Together, these results demonstrate the effectiveness of our PAG approach.
\vspace{-4mm}
\section{Conclusion}
\vspace{-4mm}
In this work, we investigated building a generalizable grasping VLA model with large-scale synthetic data. First, we curated a billion-scale grasping dataset in simulation, featuring extensive randomization and photorealistic rendering. Second, we carefully designed our model to effectively learn from synthetic action data and action-free Internet grounding data, achieving strong generalizability for grasping novel-category objects in unseen environments.
Extensive ablation studies and comparisons demonstrate that our method achieves state-of-the-art performance in table-top grasping. Furthermore, we observed that our model scales effectively with the amount of synthetic training data. Finally, we showcase that GraspVLA can acquire new grasping behaviors through few-shot post-training, highlighting its adaptability and potential for real-world applications.

\section{Limitations and Future Work}
Currently, our data generation and evaluation are conducted exclusively on the Franka Panda arm with front and side views. However, our simulation pipeline is inherently scalable and can be readily adapted to other robots and camera configurations. We leave this engineering effort as future work.

GraspVLA struggles with ambiguous instructions such as ``pick up food'' and ``pick up the leftmost object''. Addressing these challenges may require scaling vision-language pretraining and exploring architectural innovations to enhance semantic reasoning.

Like most grasping policies, we synthesize grasp labels using force-closure, which do not account for deformability—a limitation common to all such methods. Despite this, our model can still grasp certain deformable objects if their initial geometry contains convex regions enabling force closure. While previous works \cite{chen2025foldnetlearninggeneralizableclosedloop} have shown that soft-body simulation can be used to train sim2real deformable manipulation policies, we leave the integration as future work.

While current model focuses on grasping, the model design is not tailored to this specific task. We plan to extend the data generation pipeline to support other manipulation tasks, such as pick-and-place and pushing. Beyond the current modular-based expert policy used in data generation, we will explore reinforcement learning for more complex tasks like non-prehensile manipulation.

Although our PAG mechanism enables open-vocabulary grasping, it introduces additional latency.
We currently achieve around 200ms latency on NVIDIA L40s utilizing Torch Compile \cite{torchcompile}. While this is sufficient for static scenes, it may not be enough for dynamic environments, e.g., fast moving objects. Distillation and quantization techniques can be further explored.

\textbf{Acknowledgments}

This work was supported in part by National Key R\&D Program of China 62306016 and National Key R\&D Program of China 2022ZD0160201. Additionally, we extend our sincere gratitude to all our colleagues at Galbot for their assistance in collecting and annotating the post-training data.

% The acknowledgments are automatically included only in the final and preprint versions of the paper.
% \acknowledgments{If a paper is accepted, the final camera-ready version will (and probably should) include acknowledgments. All acknowledgments go at the end of the paper, including thanks to reviewers who gave useful comments, to colleagues who contributed to the ideas, and to funding agencies and corporate sponsors that provided financial support.}

\clearpage

\bibliography{references}  % .bib

\clearpage
\appendix

\section{Overview}
In the supplementary materials, we provide details about SynGrasp-1B dataset in Section \ref{supp:statistics} and \ref{supp:data-gen}. We also show that GraspVLA supports fast adaptation to new robotic arms and camera configurations in Section \ref{supp:arm_camera}. We provide details about our main experiments in Section \ref{supp:main_exp}. We also provide additional scaling law experiments in Section \ref{supp:scaling}. Details about LIBERO benchmark is provided in Section \ref{supp:libero}. Comprehensive comparison with AnyGrasp is provided in Section \ref{supp:anygrasp}. We ablate the number of camera views in Section \ref{supp:camera_view}. We also analyze the sim2real gap in Section \ref{supp:sim2real}. We provide details about the inference delay in Section \ref{supp:delay}. The implementation details of our non-blocking controller is provided in Section \ref{supp:non-blocking-ctrl}. We also provide details about the failure cases in Section \ref{supp:failure}.

\section{Details about SynGrasp-1B}
\label{supp:statistics}
\vspace{-4mm}

We present the statistics of our synthetic dataset, SynGrasp-1B, in Table~\ref{table:syngrasp}. The dataset consists of 10 million trajectories, each containing approximately 100 frames, resulting in a total of 1 billion frames—a substantial increase in scale compared to existing open-source datasets. Our dataset encompasses a diverse array of object categories, featuring 10,680 objects across 240 categories. While real-world datasets often encounter challenges related to scene diversity and require collaboration among laboratories across different countries, synthetic data generation enables us to easily create varied scenes by altering the textures of the table, ground, and walls. We utilize around 1,000 different textures for the table and 1,200 for the ground and walls, leading to a total of 1 million unique scenes.

Unlike existing datasets, SynGrasp-1B is the first to offer precise and fine-grained annotations for camera calibration, bounding box annotations, and the 3D poses of both the target object and the gripper. Thanks to our simulation engine, we can effortlessly obtain these annotations and incorporate additional types, such as depth maps and segmentation masks, when necessary. This flexibility is a significant advantage of synthetic datasets over their real-world counterparts.

\begin{table}[htbp]
\centering
\resizebox{0.8\columnwidth}{!}{
\begin{tabular}{lcccccc}
    \toprule
        & Trajectories & Objects & Scenes & \makecell{Camera \\ Calibration} & \makecell{Bbox \\ Annotation} & \makecell{3D Pose \\ Annotation} \\
    \midrule
    RoboSet \cite{bharadhwaj2023roboagent} & 98k & $<$ 200 & 11 & \xmark & \xmark & \xmark \\
    BridgeData V2 \cite{walke2024bridgedatav2datasetrobot} & 60k & 100 & 24 & \xmark & \xmark & \xmark \\
    RT-1 \cite{rt-1} & 130k & $<$ 200 & 2 & \xmark & \xmark & \xmark \\
    DROID \cite{droid} & 76k & $<$ 200 & 2080 & \checkmark & \xmark & \xmark \\
    AgiBot World \cite{agibot} & 1M & 3k & 106 & \checkmark & \xmark & \xmark \\
    \textcolor{gray}{Open-X Embodiment \cite{oxe}} & \textcolor{gray}{1.4M} & \textcolor{gray}{-} & \textcolor{gray}{311} & \textcolor{gray}{\xmark} & \textcolor{gray}{\xmark} & \textcolor{gray}{\xmark} \\
    \midrule
    SynGrasp-1B & 10M & 10k & 10M & \checkmark & \checkmark & \checkmark \\
    \bottomrule
\end{tabular}
}
\caption{\textbf{Comparison of SynGrasp-1B with Existing Datasets for Robot Manipulation.} This table highlights the advantages of SynGrasp-1B in terms of scale, annotations, and scene diversity compared to other datasets.} 
\label{table:syngrasp}
\vspace{-4mm}
\end{table}

The generation of simulation data is also much more cost-effective than real-world data collection, considering factors such as time, financial resources, space, robotic equipment, and human labor:
\begin{itemize}
    \item \textbf{Time:} A single human operator can collect only around 1,000 trajectories per day. In contrast, we can generate 10 million trajectories in 10 days using 160 NVIDIA 4090 GPUs. This efficiency accelerates the data-model feedback loop, enabling rapid model iteration and performance improvements.
    \item \textbf{Space:} Real-world data collection often necessitates multiple laboratories across different countries to enhance scene diversity. Additionally, it requires significant physical space to accommodate robots and objects; for example, the AgiBot World dataset \cite{agibot} utilizes a 4,000 $m^2$ area for data collection. In contrast, our synthetic approach does not require any physical space.
    \item \textbf{Robots:} Each human operator in real-world collection needs a physical robot, resulting in high costs and maintenance overhead.
    \item \textbf{Money:} The total cost of generating SynGrasp-1B is around \$5,000—orders of magnitude cheaper than real-world alternatives.
\end{itemize}

As an initial step toward large-scale synthetic action pre-training, we focus on grasping tasks to enable detailed analysis. Our pipeline can be extended to other robotic arms (for cross-embodiment transfer) or tasks (e.g., placing, pushing, stacking), as well as large-scale camera randomization. We leave these extensions to future work.

\begin{figure*}[htbp!]
\centering
\includegraphics[width=0.95\linewidth]{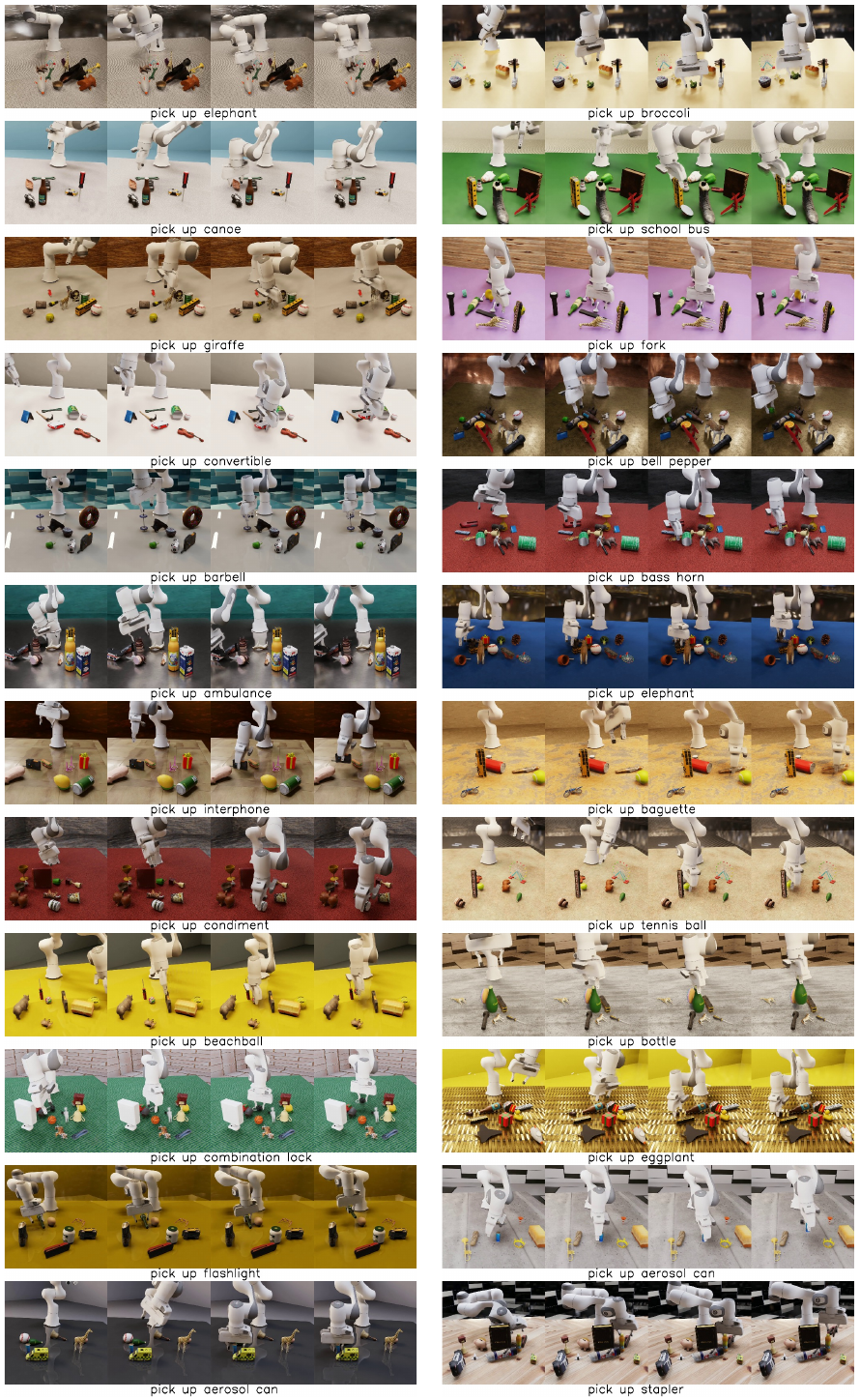}
\caption{\textbf{Gallery of SynGrasp-1B.} 24 randomly sampled trajectories from our synthetic grasping data. For clarity, 4 frames are uniformly sampled from each trajectory for display.}
\label{fig:gallery}
\end{figure*}

\textbf{Gallery.} We randomly sample 24 trajectories from our SynGrasp-1B and visualize them in Fig.~\ref{fig:gallery}. Each trajectory consists of around 100 frames, and we uniformly sample 4 frames from each trajectory for visualization.

\vspace{-4mm}
\section{Details about Data Generation}
\label{supp:data-gen}
\vspace{-4mm}

\begin{figure}[htbp!]
\centering
\includegraphics[width=0.8\linewidth]{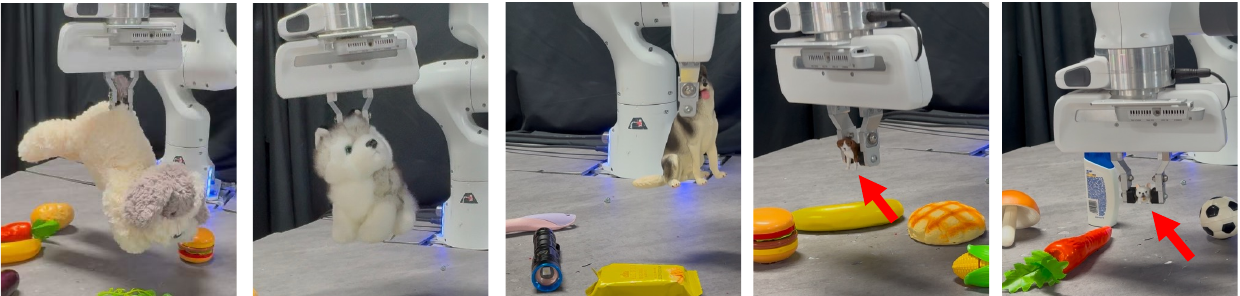}
\caption{GraspVLA handles object with diverse scales, from 2cm to 35cm.}
\label{fig:object_size}
\end{figure}

\textbf{Object Processing and Layout Generation.} 
To ensure that the scales of synthetic objects align with real-world counterparts and are suitable for grasping, we manually define minimum and maximum size constraints for each category. This helps our model to generalize to real-world objects with diverse scales. As shown in Fig. \ref{fig:object_size}, GraspVLA can grasps all scales of dog, ranging from 2cm to 35cm.
Furthermore, we simplify the object meshes using the ACVD algorithm \cite{acvd} to improve the simulation efficiency. Additionally, we randomize the height of the table, ranging from -0.1 m to 0.2 m in the robot frame.

To enhance data diversity, we create different clutter layouts for each episode by randomly placing objects within a 0.4m by 0.5m area on a table. Objects are dropped in various poses to generate physically plausible scenes. For categories requiring specific orientations, such as cups, we manually define valid poses (e.g., upright). 

The cameras are randomized within a 15 cm radius ball and rotated ±5° around each axis. Details are shown in Table \ref{tab:camera_params}.

\begin{table}[htbp]
    \centering
    \caption{\label{tab:camera_params}Camera parameters}
    \begin{tabular}{ccc}
    \toprule[2pt]
     & Position & Lookat \\ \hline
    Front Camera & x=1.35, y=0.0, z=0.54 & x=0.2, y=0.0, z=0.0 \\
    Side Camera & x=0.5, y=0.69, z=0.50 & x=0.5, y=0.0, z=0.1 \\
    \bottomrule[2pt]
    \end{tabular}
\end{table}

\textbf{Asynchronous and Grouped Data Writing.}
We employ DeepMind EnvLogger with TFDS backend \cite{noauthor_google-deepmindenvlogger_2025} for the storage of our synthetic data. Despite its clean design, considerations for high-performance large-scale simulation are necessary. Firstly, to mask the time cost of image encoding and data writing, we modified the EnvLogger implementation to perform the actual data writing operation asynchronously. Second, to avoid contention on the dataset metadata across parallel processes and to minimize data loss caused by unforeseen errors (e.g., GPU failures), the processes should not write to a single shared folder. However, if each simulation instance utilizes a unique subfolder, it results in a large number of subfolders and metadata, leading to substantial overheads in data management, transfer, and loading. As a compromise, we assign each process a subfolder with random UUIDs \cite{noauthor_uuid_2025}, and write all the trajectories within a process to the same subfolder.

\textbf{Handling Data Corruption.}
During billion-scale data generation, a simulation process could hang and get killed due to hardware faults or excessive memory consumption, resulting in file corruption or loss.
We handle these issues by proactively managing exceptions when loading the dataset with TFDS.
Upon encountering a NotFoundError, we create an empty file at the expected file path, otherwise TFDS cannot continue loading the remaining records within the subfolder.
On DataLossError, we count it as one missing record, and log the missing rate as a critical statistics for data validity.
The missing rate was below 1\%.
FailedPreconditionError is raised when the successfully loaded number of records is smaller than that in the metadata of the subfolder, and can be safely converted to a StopIteration to facilitate the correct functioning of the data loader.
The above handling of these exceptions ensures loading all the valid records.
We believe these insights will benefit future large-scale dataset efforts.

\section{Fast Adaptation to New Robotic Arm and Camera}
\label{supp:arm_camera}
\vspace{-4mm}

While we focus on specific setups for in-depth analysis, GraspVLA is not specialized for this. Our model can easily adapt to new robotic arms, grippers, and camera configurations with 5k additional synthetic trajectories (generated in one day on an NVIDIA 4090 GPU). We provide the following two examples:
\begin{itemize}
    \item \textbf{New robotic arm and gripper.} We use a UR5e arm with a Robotiq 2F-85 gripper. Since no real-world hardware is available, we test our model on the simulation environment.
    \item \textbf{New camera configuration.} We use front view and wrist view cameras. We conduct real-world experiments by mounting the camera on the wrist of the Franka Panda arm.
\end{itemize}

\begin{figure}[htbp!]
    \centering
    \includegraphics[width=0.8\linewidth]{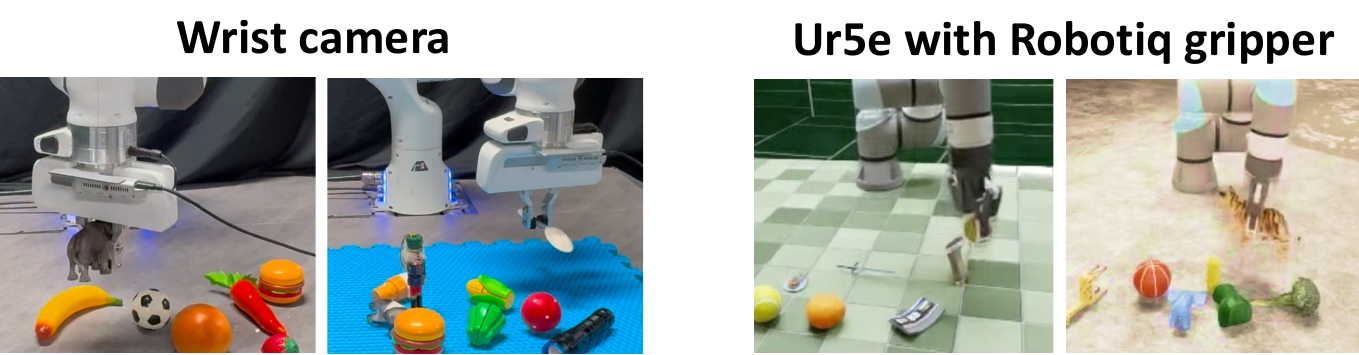}
    \caption{GraspVLA supports fast adaptation to new robotic arms and camera configurations.}
    \label{fig:new_setup}
\end{figure}

As shown in Fig. \ref{fig:new_setup} and Table \ref{tab:new_setup}, our model shows strong performance with minimal fine-tuning, enabling rapid deployment on new setups.

\begin{table}[htbp!]
    \centering
    \begin{tabular}{ccc}
        \hline
         & Wrist camera & UR5e arm with Robotiq gripper\\ \hline
        Success Rate & 76.6 & 82.1 \\ 
        \hline
    \end{tabular}
    \caption{Success rate of GraspVLA on new robotic arms and camera configurations.}
    \label{tab:new_setup}
\end{table}

\section{Details of Main Experiments}
\label{supp:main_exp}
\vspace{-4mm}

\textbf{Metrics.} In each trial, the model is allowed to attempt to grasp up to three times, with each attempt counted by the gripper closure action. Success is strictly defined as the specified object being lifted a minimum of 15 cm. The scene is not reset during each trial, even if the model knocks the object off the table. 

Additionally, we introduce Success weighted by Path Length (SPL) to further account for the number of actions taken, which is a common metric in discrete navigation tasks to evaluate the efficiency of the model. While for discrete navigation tasks, the shortest path is easy to define, for grasping tasks, the shortest path is not well defined. Therefore, for each trial, if there are several methods that can successfully grasp the object, we define the shortest path as the one with the least number of action steps. If all methods fail to grasp the object, they all get zero SPL in this trial. Note that, our SynGrasp-1B dataset stores actions in 10 Hz and all methods are trained on this dataset, so the number of action steps is comparable across methods.

% \textbf{Details about baselines.}
% We compare GraspVLA with multiple baselines including both VLA generalists and imitation learning specialists.

% \textit{OpenVLA \cite{openvla}} is an open source VLA generalist with 7 billion parameters.
% It is built upon PrimaticVLM and pre-trained with OXE and DROID dataset. 
% With its strong generalizability, it can be effectively finetuned for new settings. 

% \textit{Octo \cite{octo}}, another open-source transformer-based generalist robot policy, is pre-trained on OXE dataset, featuring efficient adaptation to new observation and action space through fine-tuning.
% We use the same training data to fine-tune Octo for grasping as we used for OpenVLA.

% \textit{Diffusion Policy\cite{dp}} is a strong baseline for visual-conditioned imitation learning, utilizing a diffusion-based action decoder for multimodal action distributions. As it lacks language conditioning, we train and test it using only the elephant category for synthetic cases, excluding web categories. For a fair comparison, we replace all objects in the standard test set with elephants and report the success rate.

% parameter selection:
%   ours: consistently good after 120k
%   pi0: largest batch size fitting a single node (paper and repo no value for pre-training), other values from paper
%  openvla: following their paper's suggestion, use LoRA fine-tune due to resource budget. with recommended parameters and default params in their github repo
%  octo: parameters from their code repo
%  dp: code repo. Hypothesis: distractors, spatial generalization

\begin{table}[htbp]
\centering
\caption{\label{tab:hyperparams}Hyperparameters for all the methods}
\begin{tabular}{c|c|c}
\toprule[2pt]
Baseline & Hyperparameter & Value \\ \hline
GraspVLA & batch\_size & 384 \\
     & learning\_rate & 1.6e-4 \\
     % & num\_steps & 120000 \\
\midrule
$\pi_0$ & batch\_size & 256 \\
        & learning\_rate & cosine schedule \\
        & warmup\_steps & 1000 \\
        & peak\_lr & 2.5e-5 \\
        & decay\_lr & 2.5e-6 \\
        & decay\_steps & 30000 \\
        % & num\_steps & 200000 \\
\midrule
OpenVLA     &     lora\_rank       &  32     \\
         &          batch\_size      &  12     \\
         & learning\_rate & 5e-4 \\
         & image\_aug & true \\
\midrule 
Octo & batch\_size & 256 \\
     & learning\_rate & rsqrt schedule \\
     & warmup\_steps & 2000 \\
     & init\_value & 0.0 \\
     & peak\_value & 3e-4 \\
     % & num\_steps & 100000 \\
\midrule 
Diffusion Policy & batch\_size & 256 \\
     & learning\_rate & cosine schedule \\
     & warmup\_steps & 500 \\
     & peak\_lr & 1e-4 \\
     & weight\_decay & 1e-6 \\
     % & num\_steps & 4000 \\
\bottomrule[2pt]
\end{tabular}
\end{table}

\textbf{Baselines.}
As Diffusion Policy does not support language conditioning, we train it using the subset of SynGrasp-1B that grasps the elephant (around 40k trajectories), and replace the target object with the elephant in the real-world experiments.
We train all other models with the full SynGrasp-1B dataset.
We train $\pi_0$ \cite{pi_0} with its pre-trained weights initialization and PaliGemma initialization for comparison.
Since OpenVLA \cite{openvla} takes a single RGB image for visual observation, we use the front camera view for it.
For Diffusion Policy \cite{dp}, we train the UNet-based version as recommended by the original paper.
For Octo \cite{octo}, we finetune the pre-trained \textit{octo-base-1.5} model.
All the models are trained with action chunks of 4 \cite{act}, except for OpenVLA, which does not support action chunking.
We provide hyperparameters of training/finetuning GraspVLA and baselines in Table \ref{tab:hyperparams}.
We run automatic evaluation in our simulation pipeline for all the baselines, continue training until the success rate converges, and select the best-performing checkpoint in the real world.
We found GraspVLA achieves a consistently high success rate after 120k steps.

\textbf{Real world setup and modification to robot finger.}
For perception, we employ an Intel RealSense D435 as the front-facing camera and a D415i as the side-facing camera. Both cameras are positioned at the center of the randomization range used in the synthetic data generation. The workspace for test objects is confined to a 40 cm x 50 cm x 20 cm area in front of the robot.

The original Franka Panda finger is too short to firmly grasp convex-shaped objects (e.g., a bottle lying on its side). This is because the hand plank collides with the top of the object, preventing the fingers from reaching deep enough to secure a stable grip. To address this issue, we extended the fingers by 2 cm in both synthetic data generation and real-world experiments.

\textbf{Details about PAG-3D.}
For steps before the gripper closure, we use the open-loop grasp pose of this trajectory as supervision. For steps after the gripper closure, we use the next step's end-effector pose as supervision.
\section{Detailed Scaling Law}
\label{supp:scaling}
\textbf{Simulation evaluation.} While the main paper analyzes the scaling law in real-world, we extend this analysis to simulation environments. Our results show that simulation is an effective proxy for predicting real-world performance. For these experiments, we use simulation environments with identical camera and table configurations to our data generation setup, but employ different object instances and materials to assess generalizability.

\begin{figure}[htbp]
\centering
\includegraphics[width=0.98\linewidth]{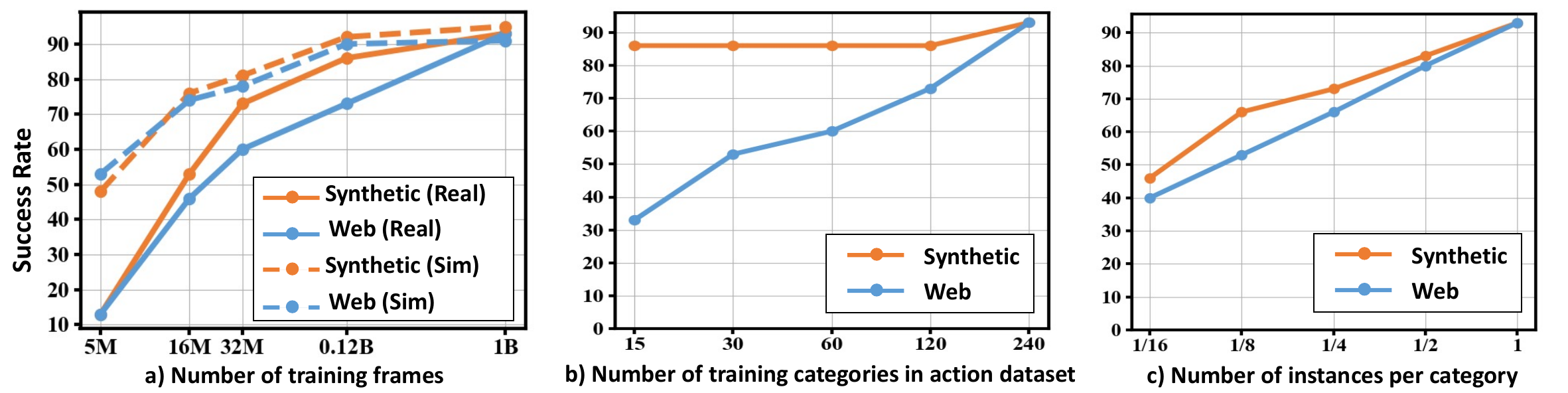}
\caption{\textbf{Scaling laws different training regimes.} (a) Performance scaling with number of training frames in both simulation and real-world environments. (b) Impact of training category diversity while fixing instances per category. (c) Effect of varying instances per category while maintaining total category count.}
\label{fig:scaling_all}
\end{figure}

As shown in Fig. \ref{fig:scaling_all}a), GraspVLA's performance on simulation data follows a scaling trend similar to that of real-world data, confirming the simulation's effectiveness for predicting real-world performance. However, we observe two key differences: (1) real-world performance scales more slowly (0.12B) compared to simulation, where performance saturates earlier, and (2) the sim-to-real gap decreases with more training frames, suggesting that larger datasets enable more robust representations and better transfer to real-world scenarios.

We further investigate how data diversity affects GraspVLA's performance by analyzing two additional scaling factors: (1) the number of training categories and (2) the number of instances per category. For each analysis, we hold the other factor and total training frames constant.

\textbf{Number of Training Categories (Fig. \ref{fig:scaling_all}b).} When varying the number of categories while fixing instances per category and total frames, performance on web categories improves steadily with more training categories, whereas performance on synthetic categories saturates early. This implies that inter-category generalization (adapting to unseen categories) benefits significantly from broader categorical coverage, while intra-category generalization (recognizing diverse instances of known categories) requires less diversity.

\textbf{Number of Instances per Category (Fig. \ref{fig:scaling_all}c).} With a fixed category count and total frames, increasing instances per category leads to consistent improvements across both synthetic and web categories. This underscores the importance of instance diversity within categories for robust generalization.
\section{Details about Experiments on LIBERO Benchmark}
\label{supp:libero}
\textbf{Setup.} 
We consider a trial successful if the robot successfully grasps and lifts the target object to a height of 10 cm. Since our model is trained with two camera views, we modify the original camera configurations provided by the LIBERO benchmark to match our training setup, aligning the camera poses accordingly. As the basket in some tasks occludes the side view severely, we remove it. Additionally, we extend the gripper by 2 cm, as detailed in the robot finger modification in Section \ref{supp:main_exp}. These adjustments are made exclusively for evaluating our model to ensure they do not affect the fine-tuned baselines, which are evaluated using the original camera configurations and gripper length.

\begin{figure}[htbp]
\centering
\includegraphics[width=0.8\linewidth]{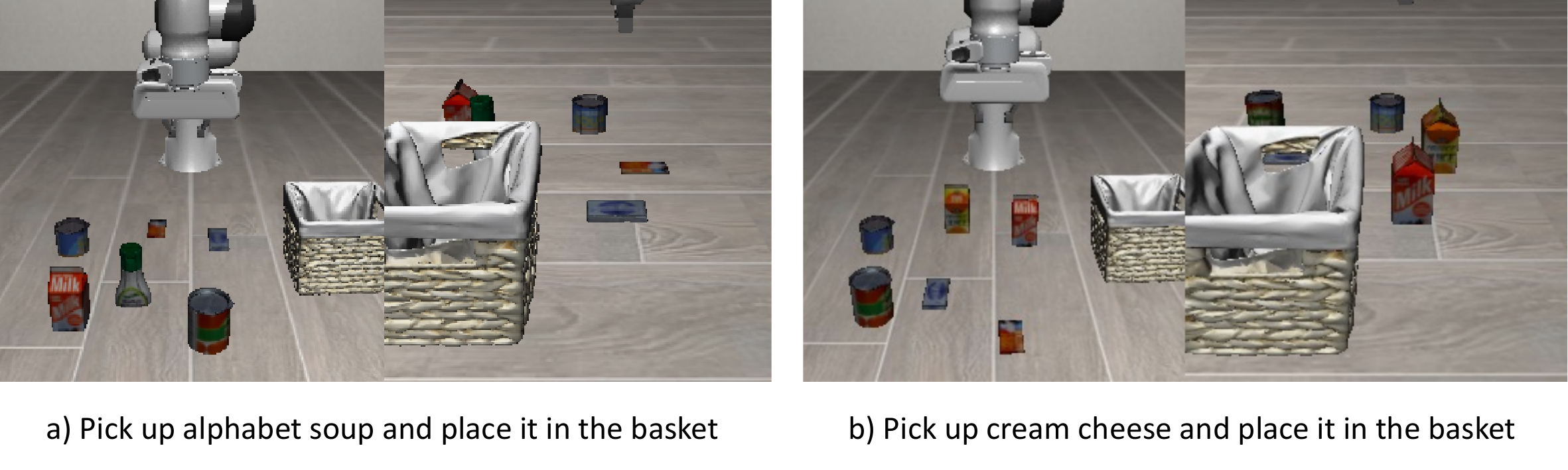}
\caption{\textbf{Examples of LIBERO Benchmark.} We visualize both front and side views side by side.}
\label{fig:libero_objects}
\end{figure}

The LIBERO-object test set presents a significant challenge for zero-shot models due to ambiguous target object descriptions. As illustrated in Figure \ref{fig:libero_objects}, even humans may struggle to identify objects like "alphabet soup" and "cream cheese" in the given scene. To account for this, we relax the success criteria: a trial is deemed successful if the robot grasps any object belonging to the same category as the target. For instance, if the target is "alphabet soup," grasping any object in the "can" category is considered a success. Similarly, if the target is "cream cheese," grasping any object in the "box" category qualifies as success.

As noted in the main paper, we exclude non-prehensile tasks to focus solely on grasping capability. We also omit tasks requiring color-based distinctions (e.g., "pick up the yellow and white mug"), as reasoning about color falls outside our scope. The specific tasks deemed invalid in the original test set, along with the modified instructions, are detailed in Tables \ref{table:libero_goal}, \ref{table:libero_object}, and \ref{table:libero_long}.

\begin{table}[htbp]
\centering
\caption{\label{table:libero_goal} \textbf{Modification to LIBERO-Goal test set.}}
\resizebox{0.7\columnwidth}{!}{
\begin{tabular}{lcc}
\toprule[2pt]
Original Caption & Valid & Modified Caption \\ \hline
put the wine bottle on top of the cabinet & \checkmark & pick up the wine bottle \\
open the top drawer and put the bowl inside & \checkmark & pick up the bowl \\
turn on the stove & x & - \\
put the bowl on top of the cabinet & \checkmark & pick up the bowl\\
put the bowl on the plate & \checkmark & pick up the bowl\\
put the wine bottle on the rack & \checkmark & pick up the wine bottle \\
put the cream cheese in the bowl & \checkmark & pick up the cream cheese box \\
open the middle drawer of the cabinet & x & - \\
push the plate to the front of the stove & x & - \\
put the bowl on the stove & \checkmark & pick up the bowl\\
\bottomrule[2pt]
\end{tabular}
}
\end{table}

\begin{table}[htbp]
\centering
\caption{\label{table:libero_object} \textbf{Modification to LIBERO-Object test set.}}
\resizebox{0.85\columnwidth}{!}{
\begin{tabular}{lcc}
\toprule[2pt]
Original Caption & Valid & Modified Caption \\ \hline
pick up the alphabet soup and place it in the basket & \checkmark & pick up the alphabet soup can \\
pick up the cream cheese and place it in the basket & \checkmark & pick up the cream cheese box \\
pick up the milk and place it in the basket & \checkmark & pick up the milk \\
pick up the tomato sauce and place it in the basket & \checkmark & pick up the tomato sauce can\\
pick up the butter and place it in the basket & \checkmark & pick up the butter box\\
pick up the orange juice and place it in the basket & \checkmark & pick up the orange juice \\
pick up the chocolate pudding and place it in the basket & \checkmark & pick up the chocolate pudding box \\
pick up the bbq sauce and place it in the basket & \checkmark & pick up the bbq sauce bottle \\
pick up the ketchup and place it in the basket & \checkmark & pick up the ketchup bottle\\
pick up the salad dressing and place it in the basket & \checkmark & pick up the salad dressing bottle\\
\bottomrule[2pt]
\end{tabular}
}
\end{table}

\begin{table}[htbp]
\centering
\caption{\label{table:libero_long} \textbf{Modification to LIBERO-Long test set.}}
\resizebox{0.98\columnwidth}{!}{
\begin{tabular}{lcc}
\toprule[2pt]
Original Caption & Valid & Modified Caption \\ \hline
turn on the stove and put the moka pot on it & \checkmark & pick up the moka pot \\
put the black bowl in the bottom drawer of the cabinet and close it & \checkmark & pick up the black bowl \\
put the yellow and white mug in the microwave and close it & x & - \\
put both moka pots on the stove & \checkmark & pick up the moka pot\\
put both the alphabet soup and the cream cheese box in the basket & \checkmark & pick up the alphabet soup can \\
put both the alphabet soup and the tomato sauce in the basket & \checkmark & pick up the alphabet soup can \\
put both the cream cheese box and the butter in the basket & \checkmark & pick up the cream cheese box \\
put the white mug on the left plate and put the yellow and white mug on the right plate & x & - \\
put the white mug on the plate and put the chocolate pudding to the right of the plate & x & - \\
pick up the book and place it in the back compartment of the caddy & \checkmark & pick up the book\\
\bottomrule[2pt]
\end{tabular}
}
\vspace{-5mm}
\end{table}

\textbf{Baselines.} For baseline models, OpenVLA and $\pi_0$ \cite{openvla, pi_0}, we use the authors' official fine-tuned checkpoints. Both models are fine-tuned on the LIBERO demonstration dataset, processed by OpenVLA to exclude static frames and failure trajectories, and rendered in high resolution.

\textbf{Impact of instruction format.} 
While the original test sets in LEBERO mainly compose of two steps, picking up an object and placing it in a container, we focus on the first step to exclusively evaluate the grasping capabilities. Therefore, to ensure a fair comparison, we also simplify the original instruction format ``pick up a {object} and place it in a {container}'' to ``pick up a {object}'' for the same instructions across all models. Note that, the instructions in the fine-tuning set are not simplified due to difficulties in segmenting and removing actions related to placing objects in containers.

As shown in Table \ref{table:supp_libero}, the performance of both fine-tuned baselines drops significantly when the instruction format is simplified. This indicates that the models are not robust to instruction variations and are sensitive to the specific instruction format. Additionally, our zero-shot model, GraspVLA, outperforms the fine-tuned models in the simplified instruction format and achieves comparable performance in the original instruction format. This demonstrates the robustness of our model to generalize to unseen environments, even in the absence of fine-tuning.

\begin{table}[htbp]
\centering
\begin{tabular}{lccc}
\toprule
 & \textbf{Long} & \textbf{Goal} & \textbf{Object} \\
\cmidrule(lr){2-4}
\rowcolor[gray]{0.85}
\textit{Format: pick up \{object\} and place it in \{container\}} & & &  \\
\rowcolor[gray]{0.85}
\textit{OpenVLA (fine-tuned)} & \textit{70.9} & \textit{78.6} & \textit{91.2} \\
\rowcolor[gray]{0.85}
\textit{$\pi_0$ (fine-tuned)} & \textbf{\textit{88.7}} & \textbf{\textit{95.4}} & \textbf{\textit{98.4}}\\ 

\midrule
\multicolumn{4}{l}{\textit{Format: pick up \{object\}}} \\
OpenVLA (fine-tuned) & 33.7 & 56.6 & 65.4 \\
$\pi_0$ (fine-tuned) & 62.7 & 79.4 & 93.8\\ 
Ours (zero-shot) & \textbf{82.0} & \textbf{91.2} & \textbf{94.1}\\
\bottomrule
\end{tabular}
\caption{\textbf{Impact of instruction format.} Fine-tuned baselines exhibit performance drops when the original instructions are simplified.} 
\label{table:supp_libero}
\vspace{-5mm}
\end{table}
\section{Details about Comparison with AnyGrasp}
\label{supp:anygrasp}
\textbf{Setup.} To ensure a fair comparison, we run the AnyGrasp baseline with up to three attempts per trial, counting it as a success if the object is grasped in any attempt. The baseline is implemented using the authors' official SDK. For perception, we use the same Franka Emika Panda robot and a RealSense D435i camera mounted on the end-effector, with the camera calibrated for accurate depth perception. Inference speed is evaluated on an NVIDIA RTX 3090 GPU.

For the language-conditioned test set, we integrate Grounding DINO \cite{liu2024groundingdinomarryingdino} to parse language instructions into bounding boxes. Grasp candidates whose 2D projections fall outside these boxes are filtered out. Given the sparse layout, this simple approach effectively eliminates irrelevant grasps. Motion planning is then used to generate trajectories for execution.

\begin{figure}[htbp]
\centering
\includegraphics[width=0.5\linewidth]{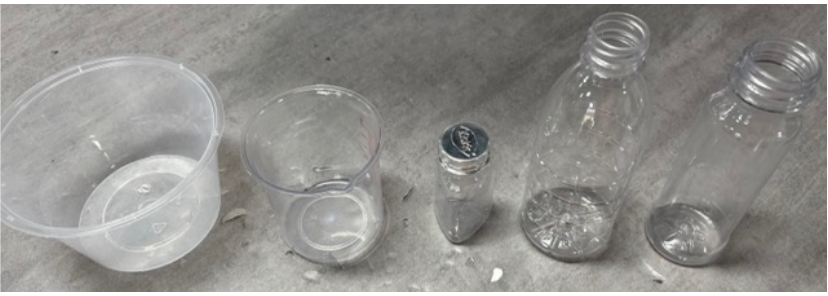}
\caption{\textbf{Transparent objects used for evaluation.}}
\label{fig:anygrasp_testset}
\vspace{-4mm}
\end{figure}

\textbf{Test Sets.} The language-driven task uses the same test set as in the main experiment (Table 1 in the main paper), comprising both synthetic and web categories for a total of 60 trials. For arbitrary grasping of common objects, we randomly select 30 objects (15 synthetic, 15 web), ensuring diffuse, non-reflective materials (e.g., rubber, wood). The transparent object test set consists of 5 objects, including 3 bottles, 1 cup, and 1 bowl. To focus on grasping transparent objects, we remove distractors from the scene and place the transparent objects at 6 different poses on the table, resulting in 6 trials per object. We visualize transparent objects in Figure \ref{fig:anygrasp_testset}.

\textbf{Analysis.} In the language-conditioned test set, the baseline fails in 5 trials. Three failures stem from incorrect bounding box predictions by Grounding DINO, largely due to ambiguities in the top-down monocular view—for example, a toy ambulance being misidentified as a charger. The remaining two failures involve flat objects (a metal fork and a plastic spoon), where the point clouds merge with the table surface, rendering the objects indistinguishable even to human observers. Transparent objects pose a similar challenge, as missing depth information leads to point-cloud-based grasping failures. However, since RGB images reliably capture these objects, our RGB-based model overcomes these limitations and succeeds where the baseline fails.

Overall, AnyGrasp and our method provide complementary solutions. AnyGrasp is a fast grasping detection model and adapting it to open-vocabulary grasping requires extra modules (segmentation, motion planner, failure recovery)—each introducing potential failures. For instance, collision-free path planning often fails in cluttered scenes like our post-train Task 3. In contrast, our model is \textbf{end-to-end}, \textbf{closed-loop}, and \textbf{easily adapts to specialized tasks} (e.g., grasping in specific poses) without requiring task-specific modules. Besides, AnyGrasp uses depth as input, which suffers from incomplete and noisy issues for transparent materials. In contrast, our model relies solely on RGBs, bypassing this issue.
\section{Ablation of Camera Views}
\label{supp:camera_view}
$\pi_0$ natively supports multi-camera, so we fine-tune it with the same front and side views as our model to ensure fair comparison. However, OpenVLA only supports single view, so we ablate the number of views here and show that our single-view version outperforms OpenVLA by 40\%. 

\textbf{Impact of the Number of Input Views.}
To ensure a fair comparison, we use only front-view images as input for our method, consistent with the single-view baseline OpenVLA. As shown in Table \ref{table:single-view}, this constraint results in approximately 30\% lower performance compared to our multi-view approach. Nevertheless, our model still achieves 40\% higher performance than OpenVLA, demonstrating the effectiveness of our design.

\begin{table}[htbp]
\centering
\resizebox{0.5\textwidth}{!}{
\begin{tabular}{c|cc}
\toprule[2pt]
Model & Synthetic & Web \\ \hline
OpenVLA (single-view) & 20.0 & 3.3 \\
Ours (single-view) & 60.0 & 56.6 \\
Ours (multi-view) & \textbf{93.3} & \textbf{93.3}\\
\bottomrule[2pt]
\end{tabular}
}
\caption{\textbf{Impact of number of input views.} Comparison of GraspVLA with different numbers of input views. The results demonstrate that while multiple views significantly improve performance, our single-view implementation still outperforms the OpenVLA baseline by 40\%.}
\label{table:single-view}
\end{table}
\section{Mitigation of Sim-to-Real Gap}
\label{supp:sim2real}
In this section, we examine the sim-to-real gap in the context of training a VLA model for grasping using imitation learning. The sim-to-real gap primarily appears in two key areas: visual appearance and physical dynamics.

\textbf{Visual appearance.} Thanks to advances in pre-trained vision encoders and ray-traced rendering, the visual discrepancy between synthetic and real-world RGB images has significantly narrowed. By leveraging diverse material and texture datasets, we can generate realistic scenes that cover a wide range of robotic grasping scenarios—far more efficiently than collecting equivalent real-world data across varied environments (as discussed in \ref{supp:statistics}). Even when certain material or texture combinations appear unrealistic (e.g., a red table against a green wall), the model still learns generalizable representations from such diversity, consistent with findings in \cite{simtoreal}. Additionally, co-training with large-scale Internet vision-language datasets further enhances the model's robustness to visual discrepancies \cite{pi0_5}.

\textbf{Physical dynamics.} The sim-to-real gap in physical dynamics arises mainly from inaccuracies in modeling material properties (e.g., surface friction), contact dynamics (e.g., forces, friction, deformations), and actuator/sensor behavior. In this work, we mitigate this gap through three key design choices:
\begin{itemize}
    \item \textbf{Simplified control.} We use positional control and treat gripper actions as discrete open/close commands, avoiding complex dynamics modeling.
    \item \textbf{Stability filtering.} We only keep grasps that forms force-closure under low friction coefficient (0.15), ensuring the model prioritizes robust strategies.
    \item \textbf{Geometry-driven planning.} We focus on mesh-based grasp poses rather than dynamics-dependent policies, enhancing robustness to physical variations.
\end{itemize}

While these strategies effectively reduce the sim-to-real gap for grasping, they may not generalize to tasks requiring fine-grained dynamics understanding, such as non-prehensile manipulation. We leave the investigation of such scenarios to future work.
\section{Inference Delay}
\label{supp:delay}

The combination of autoregression and flow matching in GraspVLA introduces additional inference delay.
Based on Section 5.6 and Table \ref{table:inference_time}, while PAG is critical for a high grasp success rate, it contributes to $\sim$ 63\% inference delay due to 14 additional tokens to generate.
We leave the further improvement of the inference efficiency with PAG as future work.
We additionally found that the prefill stage has a similar delay as the decode stage, which could be due to a low GPU utilization with single-sample inference.

\begin{table}[htbp]
\centering
\begin{tabular}{l|l}
\toprule[2pt]
\textbf{component}                 & \textbf{inference time (ms)} \\ \hline
vision encoder            & 9                   \\
bounding boxes (8 tokens) & 72                  \\
grasp pose (6 tokens)     & 50                  \\
flow matching             & 64                  \\
\bottomrule[2pt]
\end{tabular}
\caption{\textbf{Breakdown of inference time on NVIDIA L40s GPU.}}
\label{table:inference_time}
\end{table}
\vspace{-4mm}
\section{Non-Blocking Control}
\label{supp:non-blocking-ctrl}
\vspace{-4mm}
We explore the implementation of non-blocking controller for smooth action.
We implement a Cartesian-space impedance controller adapted from Franka ROS \cite{franka_ros} and SERL Franka Controllers \cite{luo2024serl}. 
The architecture converts Cartesian impedance commands into joint-space control via real-time Jacobian-based transformation with singularity handling, while optimizing impedance parameters through system identification. 

\begin{figure}[htbp]
\centering
\includegraphics[width=0.7\linewidth]{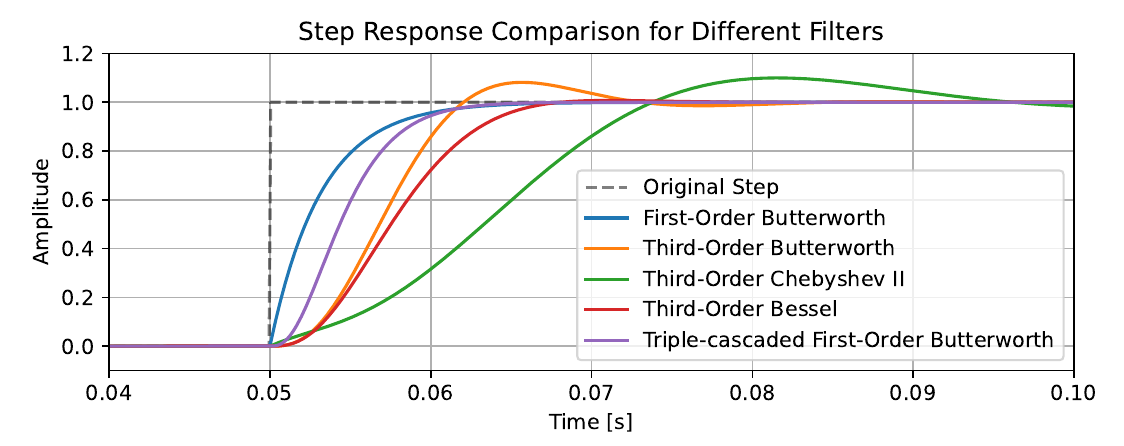}
\caption{\textbf{Step response comparison for different filters}. In the comparison, these filters set the same sampling frequency and cutoff frequency. The first-order Butterworth filter exhibits a large initial acceleration in its step response. The third-order Butterworth filter, third-order Chebyshev II filter, and third-order Bessel filter all exhibit overshoot to varying degrees. The triple-cascaded first-order Butterworth filter can avoid excessive initial acceleration and eliminate overshoot while maintaining convergence speed, making it an ideal filter choice.}
\label{fig:step_response_comparison}
\end{figure}

To mitigate abrupt target transitions, we evaluated multiple filter implementations and selected a cascaded filter design for its superior smoothing performance (Figure ~\ref{fig:step_response_comparison}). It achieves fast convergence without overshoot while avoiding excessive initial acceleration, which is suitable for the output characteristics of the GraspVLA model. Additionally, positional interpolation was adopted instead of temporal interpolation to address synchronization mismatches between model computation latency and control pipelines. 

To achieve more fluent and coherent motions, GraspVLA generates multi-step predictions at each inference cycle. These predictions are incorporated via a receding-horizon optimization scheme within the asynchronous control architecture, where filter and interpolation strategies are systematically applied to the predicted trajectory. This non-blocking control architecture proactively compensates for computational latency variations while ensuring smooth interleaving of control actions, which significantly mitigates oscillatory patterns in dynamic manipulation scenarios.

While we employ non-blocking control for demonstration recording to achieve natural trajectories, all experimental evaluations use blocking control to ensure rigorous performance measurement.
\section{Failure Analysis}
\label{supp:failure}
To thoroughly assess the limitations of our approach, we conduct a detailed failure analysis. Since the test set in the main paper reveals only two failure cases—which may not be representative—we design a significantly more challenging test set featuring cluttered scenes. Specifically, we randomly place objects across the table to cover the entire workspace and stack some objects (e.g., placing a strawberry on top of a bulldozer) to create complex, occluded scenarios. We then evaluate the model's performance under these conditions and identify the primary failure modes.

The most frequent failure case (31\%) occurs when the model hesitates due to ambiguous language instructions, such as when multiple objects match the description (e.g., two target bottles). This could be mitigated by incorporating longer contextual history. The second most common issue (27\%) arises in highly cluttered scenes, where the model misidentifies objects, likely due to insufficient training data for such scenarios. Future work could utilize advanced data augmentation methods and generative modeling techniques to create more diverse and complex training samples. Another notable failure mode (21\%) involves objects with smooth surfaces (e.g., plastic balls) slipping during grasping, which tactile feedback might help resolve. Additionally, when the target object is occluded (14\%), the model struggles to grasp it precisely, suggesting a need for active perception techniques. Finally, the remaining failures (7\%) include minor errors such as early gripper closure or collisions with the environment, which reinforcement learning could potentially address. We leave these potential improvements for future work.

% \bibliographystyle{plainnat}
% \bibliography{references}

% \clearpage

% \end{document}

\end{document}